\documentclass[journal,twoside,web]{ieeecolor}
\usepackage{tmi}
\usepackage{cite}
\usepackage{amsmath,amssymb,amsfonts}
\usepackage{graphicx}
\usepackage{textcomp}

\usepackage{algorithm}
\usepackage{algpseudocode}
\usepackage{multirow}
\usepackage{multicol}
\usepackage{bigstrut}
\usepackage{color}
\usepackage{colortbl}
\usepackage{hyperref}
\usepackage{tabularx}
\usepackage{adjustbox}
\usepackage{booktabs}
\usepackage[normalem]{ulem}
\useunder{\uline}{\ul}{}
\usepackage{marginnote}

\definecolor{red}{rgb}{0.8,0,0}
\definecolor{blue}{rgb}{0,0,0.8}
\definecolor{green}{rgb}{0,0.4,0}

\newcommand{\change}[2]{}
\newcommand{\lchange}[2]{}

\newcommand{\changed}[3]{%
  \change{#1}{#2}
  #3
}

\usepackage{etoolbox}
\makeatletter
\@ifundefined{color@begingroup}%
{\let\color@begingroup\relax
\let\color@endgroup\relax}{}%
\def\fix@ieeecolor@hbox#1{%
\hbox{\color@begingroup#1\color@endgroup}}
\patchcmd\@makecaption{\hbox}{\fix@ieeecolor@hbox}{}{\FAILED}
\patchcmd\@makecaption{\hbox}{\fix@ieeecolor@hbox}{}{\FAILED}
\def\BibTeX{{\rm B\kern-.05em{\sc i\kern-.025em b}\kern-.08em
    T\kern-.1667em\lower.7ex\hbox{E}\kern-.125emX}}
\begin{document}

% The responses to the reviewer comments
% -> see responses.tex
% \include{responses}
% \clearpage
% \twocolumn

% % \clearpage
% \setcounter{figure}{0}
% \setcounter{table}{0}
% % \addtocounter{figure}{-14}
% % \addtocounter{table}{-7}
% \pagenumbering{arabic}
% \setcounter{page}{1}

\title{PASS:Test-Time Prompting to Adapt Styles and Semantic Shapes in Medical Image Segmentation}
% \title{Interactive/Hybrid Prompt Learning for Test-Time Adaptation in Medical Image Segmentation}
% \title{Uncovering Robust Prompt for Test-Time Adaptation in Medical Image Segmentation}
\author{Chuyan Zhang, Hao Zheng, Xin You, Yefeng Zheng, \IEEEmembership{Fellow, IEEE}, and Yun Gu, \IEEEmembership{Member, IEEE}
\thanks{This work is supported in part by Shanghai Municipal of Science and Technology Project, under Grant 20JC1419500 and Grant 20DZ2220400. (Corresponding author: Yun Gu.)}
\thanks{Chuyan Zhang,  Xin You and Yun Gu are with the Institute of Medical Robotics, Shanghai Jiao Tong University, Shanghai, China. (Email: {zhangchuyan, sjtu\_youxin, geron762}@sjtu.edu.cn)}
\thanks{Hao Zheng and Yefeng Zheng are with the Jarvis Research Center, Tencent YouTu Lab, Shenzhen, China. (Email: {howzheng, yefengzheng}@tencent.com)}.
}

\maketitle

\begin{abstract}
Test-time adaptation (TTA) has emerged as a promising paradigm to handle the domain shifts at test time for medical images from different institutions without using extra training data.
However, existing TTA solutions for segmentation tasks suffer from (1) dependency on modifying the source training stage and access to source priors or (2) lack of emphasis on shape-related semantic knowledge that is crucial for segmentation tasks.
  \textcolor{black}{
  Recent research on visual prompt learning achieves source-relaxed adaptation by extended parameter space but still neglects the full utilization of semantic features, thus motivating our work on knowledge-enriched deep prompt learning.
  Beyond the general concern of image style shifts, we reveal that shape variability is another crucial factor causing the performance drop. To address this issue, we propose a TTA framework called PASS (Prompting to Adapt Styles and Semantic shapes), which jointly learns two types of prompts: the input-space prompt to reformulate the style of the test image to fit into the pretrained model and the semantic-aware prompts to bridge high-level shape discrepancy across domains.
  Instead of naively imposing a fixed prompt, we introduce an input decorator to generate the self-regulating visual prompt conditioned on the input data.
  To retrieve the knowledge representations and customize target-specific shape prompts for each test sample, we propose a cross-attention prompt modulator, which performs interaction between target representations and an enriched shape prompt bank.}
  Extensive experiments demonstrate the superior performance of PASS over state-of-the-art methods on multiple medical image segmentation datasets. The code is available at \url{https://github.com/EndoluminalSurgicalVision-IMR/PASS}.
% Keep the abstract to 250 words or less.
\end{abstract}

\begin{IEEEkeywords}
test-time adaptation, prompt learning, transfer learning
\end{IEEEkeywords}

\section{Introduction}
\label{sec:intro}
\IEEEPARstart{S}{emantic} segmentation is a critical task in the realm of medical image diagnosis, enabling the precise delineation of anatomical structures. 
Domain adaptation (DA) techniques\cite{zhao2020review, wu2021unsupervised, bateson2022source, liu2020shape, wu2023upl} have propelled deep-learning medical segmentation models into the clinical practice, overcoming the challenge of ubiquitous distribution shifts induced by diverse devices and scanning protocols. In real-world scenarios where only pretrained model weights are accessible (due to issues such as data privacy), test-time adaptation (TTA)\cite{wang2021tent,wang2022cotta, bateson2022moment,he2020self, niu2023sar} emerges as a promising alternative to DA. TTA emphasizes learning directly from few unlabeled target test data, while DA methods typically require access to source data\cite{liu2020shape, wu2021unsupervised} or extensive target training data\cite{bateson2022source, wu2023upl}.

% \begin{figure}[t!]
% \centering
% \includegraphics[width=1\linewidth]{Fig/shape_glcm_des.pdf}
% \caption{The relationship between source-to-target performance and shape/style texture similarity during adaptation in multi-site MRI prostate (one source $\rightarrow$ five target datasets: BMC/HK/BIDMC/I2CVB/UCL) and optic disc-cup (one source $\rightarrow$ three target datasets: BASE1-3) image segmentation. We use the moment descriptor defined in \cite{bateson2022moment} and gray-level co-occurrence matrix (GLCM) features\cite{haralick1973textural} to respectively represent the shape of objects and the style texture of images. The average cosine similarity of source and target shape/texture descriptors is computed. For the same source model, the transfer performance on different target datasets is highly correlated with source-target shape similarity (resembling polygons in (a)-(b)).}
% %, implying that adaptation performance degradation is partly caused by domain shape inconsistency.}
% %We can observe that the source model performs better on the target dataset that possesses similar shape descriptors, which implies that the adaptation performance degradation is partly caused by shape inconsistency. Hence, we are motivated to learn target-specific shape prompts to boost the model transferability.}
% \label{fig::shape_descriptor}
% \end{figure}
\begin{figure}[t!]
\centering
\includegraphics[width=1\linewidth]{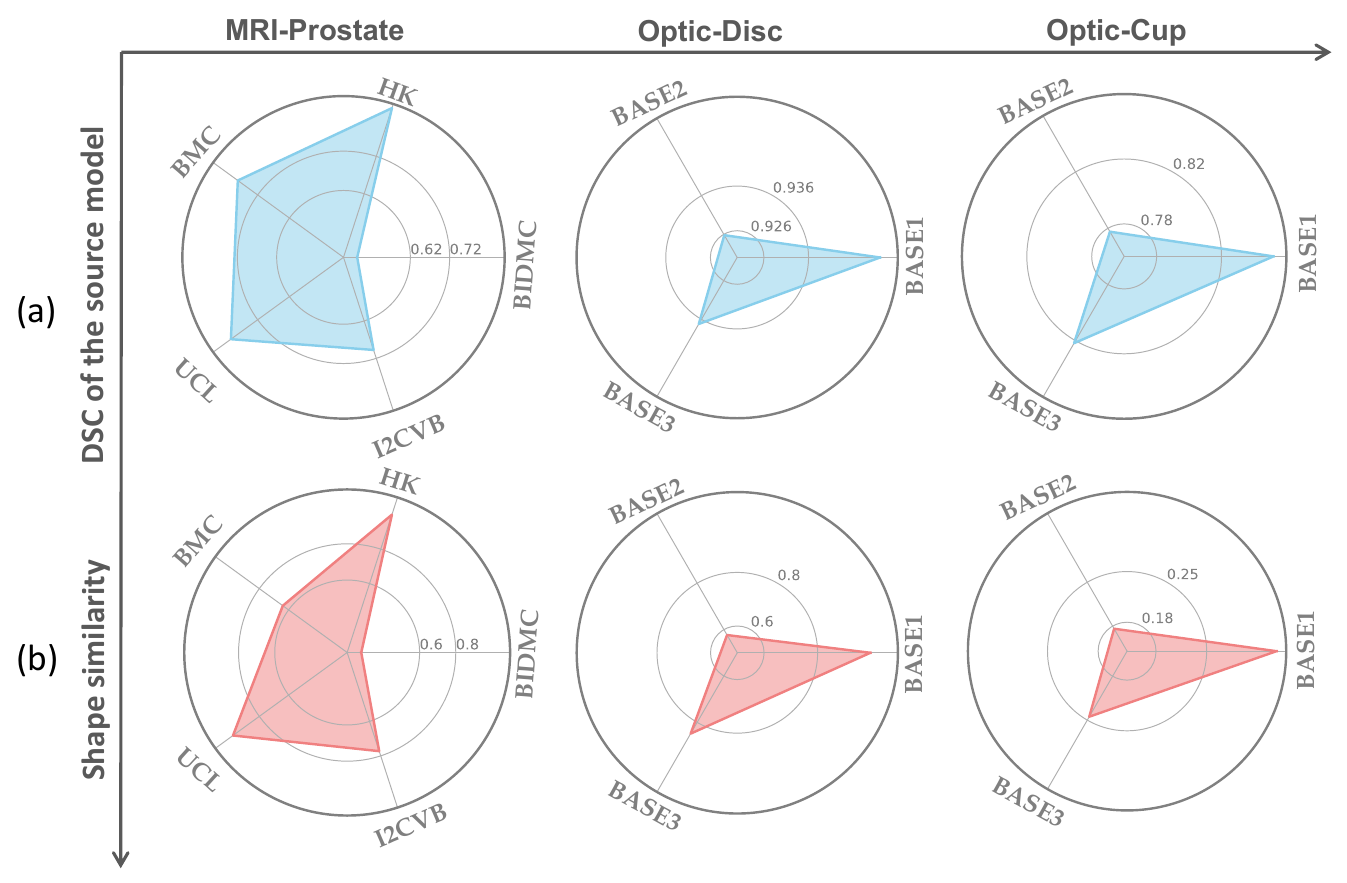}
\caption{The relationship between source-to-target performance and shape similarity during adaptation in multi-site MRI prostate (one source $\rightarrow$ five target datasets: BMC/HK/BIDMC/I2CVB/UCL) and optic disc-cup (one source $\rightarrow$ three target datasets: BASE1-3) image segmentation. We use the moment descriptor defined in \cite{bateson2022moment} to represent the shape of objects. The average cosine similarity of source and target shape descriptors is computed. For the same source model, the transfer performance on different target datasets is highly correlated with source-target shape similarity (resembling polygons in (a)-(b)).}
%, implying that adaptation performance degradation is partly caused by domain shape inconsistency.}
%We can observe that the source model performs better on the target dataset that possesses similar shape descriptors, which implies that the adaptation performance degradation is partly caused by shape inconsistency. Hence, we are motivated to learn target-specific shape prompts to boost the model transferability.}
\label{fig::shape_descriptor}
\end{figure}
A line of TTA approaches\cite{he2020self, sun2020test, valanarasu2022dpg, he2021autoencoder, guo2023atals} constructed self-supervised auxiliary branches, such as auto-encoders\cite{he2020self, valanarasu2022dpg, he2021autoencoder}, to drive the adaptation of pretrained models to target domains. These methods entail altering the pretraining paradigm. However, in practice, it is critical to isolate the source training and adaptation stages due to regulatory and privacy considerations. Several unsupervised solutions have been present to relieve the dependency on intricate source training: (1) \textbf{Model-driven TTA methods}, which harness task-agnostic knowledge from the pretrained model via entropy minimization\cite{wang2021tent, hu2021rncr, niu2023sar, wu2023upltta}, transformation invariance\cite{nguyen2023tipi, zhang2023oclttt} and model ensembling\cite{wang2022cotta, wu2023upltta}, or align domain statistics stored in Batch Normalization (BN) layers\cite{nado2020ptbn, mirza2022dua}; (2) \textbf{Source-driven TTA methods}, which use manually defined semantic-relevant priors from source data as regularisation for the adaption stage, including class-ratio prior\cite{bateson2022source} and shape prior\cite{bateson2022moment, liu2022single}. %Another line of TTA approaches \cite{wang2021tent, bateson2022moment,hu2021rncr, bateson2022moment,liu2022single, zhang2023oclttt} focus on designing unsupervised losses: entropy minimization \cite{wang2021tent, niu2023sar}, contour regularization \cite{hu2021rncr}, class-ratio prior loss \cite{bateson2022source}, shape prior loss \cite{bateson2022moment, liu2022single}, self-training loss \cite{wang2022cotta} and transformation-invariance loss \cite{nguyen2023tipi, zhang2023oclttt}. 

Drawing inspiration from the success of prompt tuning \cite{jia2022visual, zhou2022coop, shu2022test} in enabling a pretrained model to solve a new task without sacrificing previously learnt knowledge, a set of \textbf{target-driven} domain adaptation methods \cite{hu2022prosfda, chen2023vptta} have been explored to capture information from target images themselves via prepending visual prompts to the input images. They assume that differences in texture styles are accountable for cross-domain shifts and thus propose to modify the input distributions via learnable input-space prompts. These researches have shown that, beyond the proven advantages in efficiently tuning large-scale foundation models to new tasks with sufficient labelled target data\cite{jia2022visual}, prompt learning also performs well in unsupervised domain adaptation of compact models for specific semantic tasks with unseen label spaces.

\textcolor{black}{
% Previous works are mainly based on the assumption that the shape pattern of an organ is fully consistent in domain adaptation. However,
However, segmentation tasks pose unique obstacles in applying existing TTA methods, which involve maintaining the structured shape of predictions.
%It is generally recognized that style shifts in the input space across domains are the main cause of hampered model generalization while the shape pattern of an organ is fully consistent in domain adaptation.
There are often cases that deviate from the normative shape distribution in terms of position, size or appearance, which are at risk of erroneous predictions. Some descriptors have been proposed to quantify differences in shape across domains \cite{bateson2022moment}. Figure \ref{fig::shape_descriptor} implies that cross-domain shape similarity exhibits a high correlation with the transfer performance across multiple datasets. This proves that shape variance is also a substantial factor contributing to the adaptation performance degradation for segmentation tasks.
Most TTA solutions only focus on mitigating low-level domain shift in the input space\cite{hu2022prosfda, chen2023vptta} or tightening the probability manifold in the output space\cite{wang2021tent, hu2021rncr, niu2023sar, wu2023upltta, nguyen2023tipi, zhang2023oclttt, hu2022prosfda, chen2023vptta}.
The lack of explicit constraints on the enriched shape representations in the latent space of networks leads to unstable adaptations or even model collapse.
%Most of previous TTA solutions \cite{wang2021tent, hu2021rncr, niu2023sar, wu2023upltta, nguyen2023tipi, zhang2023oclttt, nado2020ptbn, mirza2022dua, hu2022prosfda, chen2023vptta} lack explicit constraints on shape representations, resulting in unstable adaptation or even model collapse. 
Additionally, some source-driven approaches \cite{bateson2022source, bateson2022moment, liu2022single} impose source shape priors to supervise the target objects. Despite the advantages of explicit prior constraints for stabilizing adaptation, data with out-of-distribution shape patterns may still lead to error accumulation.}

% In this paper, we first introduce the concept of shape prompts into target-driven TTA, called \textbf{S}hape \textbf{P}rompt learning \textbf{TTA} (SPTTA).
In this paper, we propose a TTA framework called \textbf{Prompting}
to \textbf{A}dapt \textbf{S}tyles and \textbf{S}emantic shapes (PASS) to enhance target-specific shape representations.  We first introduce the concept of shape prompts into target-driven TTA.
Specifically, we add learnable shape prompts upon the latent features to customize knowledge from the source model towards the target data. Considering the intra-domain variance, we design a cross-attention prompt modulator to assemble the most related shape prompts from an enriched domain-common prompt bank for every single test sample.
Besides, we tackle the commonly-recognized texture style shift \cite{hu2022prosfda, chen2023vptta} by constructing an input decorator that prompts the source model with the input-specific visual perturbation. 
Different from the same visual prompt for the whole domain in \cite{hu2022prosfda} and the frequency-domain prompt that focuses on a small range of low-frequency domain \cite{chen2023vptta}, the proposed input decorator dynamically generates a comprehensive visual prompt for each data.
To avoid error accumulation in the continually online TTA setting, we further propose an alternating momentum updating (AMU) strategy. The intuition behind AMU is that there should be a buffer to store historical knowledge and an online model to fit a particular sample. 
For stable adaptation, we present a momentum decay factor to ensure that the later samples have less impact on the buffer.

Our contributions are summarized as follows: 
(1) Except for the well-recognized style shift, we first highlight the issue of shape variability in the test-time adaptation of segmentation models. Without restrictions on source training or access to the source domain, we propose a TTA method named PASS To mitigate both the style and shape inconsistency across domains.
%to jointly learn the instance-aware style prompt and semantic-guided shape prompt.
(2) An input decorator is introduced to reformulate the input data via instance-aware style prompt and a cross-attention prompt modulator is proposed to selectively customize the specific shape prompt for each test sample from an enriched shape prompt bank.
Cooperatively, a novel alternating momentum updating strategy is proposed to mitigate overfitting to the online test data and improve the learning capacity of prompts. 
(3) Extensive experimental results demonstrate that our method with a few trainable parameters could achieve state-of-the-art performances in both online and offline TTA settings for multi-center optic disc/cup and magnetic resonance imaging (MRI) prostate segmentation tasks.
%For semantic segmentation, most TTA works have some limitations: (1) require dedicated auxiliary task branches, (2) ignore the complexity of segmentation tasks, such as shape inconsistency (3) necessitate extracting prior information from the source data. learn reasonable visual prompts and enhance feature generalization in a progressive manner. 
\section{Related Work}
\begin{figure*}[t]
\centering
\includegraphics[width=1\linewidth]{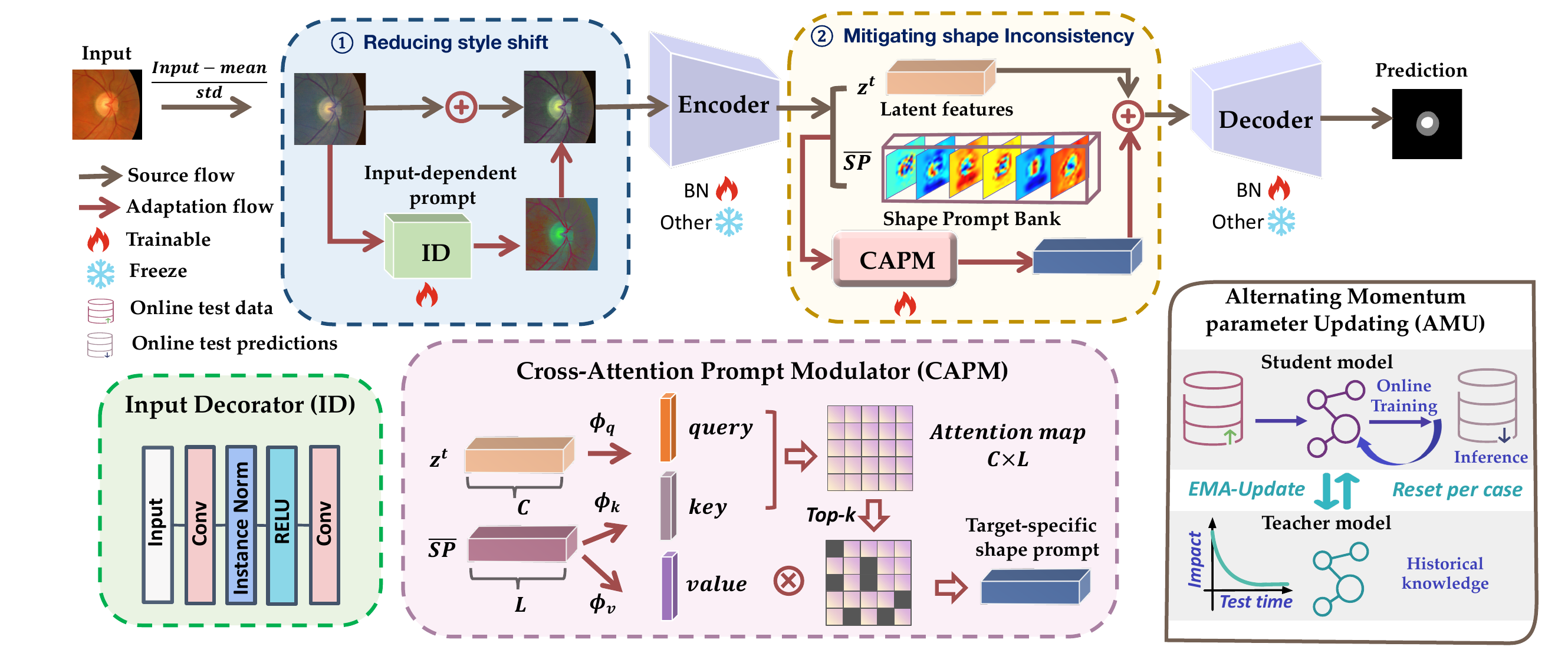}
\caption{An overview of the proposed PASS approach, comprising three components: (1) an input decorator to reduce the style shift, (2) a cross-attention prompt modulator to mitigate the shape inconsistency, 
%The former generates the input-dependent prompt to reduce the style shift across domains in the input space. The latter distills target-specific shape prompt from the enriched domain-common shape prompt bank to mitigate the shape inconsistency across domains in the latent space.
and (3) an alternating momentum parameter updating strategy with the teacher/student framework to avoid error accumulation during the adaptation stage. The weights of the teacher model are updated by the exponential moving average (EMA) with a decay factor from the online updating student model. For each new test data, the student model is reset by the teacher model to recover the historical knowledge and is adapted to the given sample.}
\label{fig::overview}
\end{figure*}
\subsection{Test-time Adaptation (TTA)}
TTA aims to reduce the performance gap when deploying a source model on another target domain by relying solely on unlabeled test data. To cope with the absence of labels, numerous works \cite{he2020self, sun2020test, valanarasu2022dpg, he2021autoencoder, guo2023atals,  karani2021dae} introduced self-supervised auxiliary tasks to drive the adaptation stage. Especially for segmentation tasks, auto-encoders are preferable for capturing fine-grained representations \cite{he2020self, valanarasu2022dpg, he2021autoencoder, karani2021dae}. However, these approaches necessitate customized pretraining stages and even extra pretrained models, deviating from the original intent of performing flexible adaptation only during test time. Another line of work seeks reasonable unsupervised loss functions. TENT-based methods \cite{wang2021tent, hu2021rncr, niu2023sar, wu2023upltta} utilize entropy-based losses to increase the confidence of model predictions. Nguyen \emph{et al.} \cite{nguyen2023tipi} and Zhang \emph{et al.}\cite{zhang2023oclttt} imposed the invariance regularization on model predictions under input transformations. Wang \emph{et al.} \cite{wang2022cotta} and Wu \emph{et al.} \cite{wu2023upltta} ensembled multiple models to reduce the uncertainty and improve the training stability. Moreover, Batch Normalization (BN) layers also provide a shortcut to align the distributions \cite{nado2020ptbn, mirza2022dua}. The aforementioned methods are task-agnostic and neglect the inherent characteristics of segmentation tasks. To maintain the structural properties of objects, some methods manually extract priors from the source data annotations to constrain the target predictions, including class-ratio prior \cite{bateson2022source} and shape prior \cite{bateson2022moment, liu2022single}. This paper focuses on studying TTA for segmentation tasks without modifying the pretraining stage and the access to source data. To further tackle the shape inconsistency that contributes to the performance drop during adaptation, our PASS learns target-specific shape information for every single test sample.
\subsection{Prompt Learning}
Prompting was first proposed in the form of text queries to instruct a large-scale language model to perform various downstream tasks \cite{jia2022visual, zhou2022coop, liu2023pre}. Recently, prompt tuning \cite{jia2022visual} has been explored to boost the generalizability of pretrained vision models. For example, VPT \cite{jia2022visual} prepends trainable visual prompts to input images when transferring the frozen pretrained backbone to a new task. The deficiency in VPT lies in its limited adaptation capability of adding the same prompts to all inputs, regardless of data diversity. To further improve the effectiveness of visual prompts, some subnetworks are built to produce dynamic prompts \cite{loedeman2022prompt, nie2023pro}. Apart from the fine-tuning paradigm, a few works attempt to introduce learnable prompts in bridging the distribution gap for domain generalization \cite{shu2022test} and domain adaptation \cite{hu2022prosfda, gan2023decorate, chen2023vptta}, where input-space prompts are added to the image/text and unsupervised losses are adopted to optimize the prompts. By contrast, we employ test-time adaptive prompts in both input space and latent space to mitigate the input distribution gap and shape inconsistency for TTA in medical image segmentation. 
%we learn adaptive prompts on the fly with a single test sample,
%This paper employs test-time prompts in both input space and latent space to perform TTA in medical image segmentation. In the input space, an input decorator generates sample-specific visual prompts, which modify the input distribution to fit in the pretrained model. In the latent space, the shape prompts provide target-relevant shape information.
\section{Methodology}
\subsection{Preliminary and Problem Definition}
\textbf{Test-time training paradigm.} In domain adaptive segmentation, let $\mathit{X}^s/\mathit{X}^t$ denote the source/target input space, and $\mathit{Y}^s/\mathit{Y}^t$ be the source/target label space.
% Given a segmentation model $\mathcal{S}_{\psi}$ pretrained on source domain(s) $\mathit{D}^S =\{(x^S, y^S)\mid x^S \in \mathit{X}^S, y^S \in \mathit{Y}^S\}$, the purpose of TTA is to adapt $\mathcal{S}_{\psi}$ on a set of unlabeled test data $\mathit{D}^T=\{x_^T: x^T \in \mathit{X}^t\}$ in another target domain so that $f_{\theta_T}$ can accurately map $x_T$ to unseen $Y_T$.  
Given a segmentation model $\mathcal{S}_{\psi^*}$ pretrained on source domain(s) $\mathit{D}^s =\{(x^s, y^s)\mid x^s \in \mathit{X}^s, y^s \in \mathit{Y}^s\}$, test-time adaptation (TTA) aims at adapting this model to perform well on a set of unlabeled test data in another target domain $\mathit{D}^t=\{x^t_i \mid x^t_i \in \mathit{X}^t, i=1,...,N^t\}$.  
Formally, following the above notations and standard TTA paradigm, we optimize the target model $\mathcal{T}$ with an unsupervised test-time training loss function $L_{tt}$:
\begin{equation}
    \mathcal{T_{\theta^*}} = \arg \min_{\theta} \mathbb{E}_{x^t\sim\mathit{D}^t}[L_{tt}(\mathcal{T}_{\theta}(x^t))]
\end{equation}

\textcolor{black}{
\textbf{Prompt learning with unlabeled test data.} In prompt learning, $\theta$ contains additional trainable parameters $\phi$ for adaptation: $\theta = \{\psi, \phi\}$, where $\psi$ is initialized by $\psi^*$ from the pretrained model $S$ and $\phi$ is randomly initialized. Except for BN parameters and the introduced parameters $\phi$ which are trainable, the rest of the parameters are usually frozen during adaptation. The objective loss function $L_{tt}$ for test-time training is always carefully constructed in a self-supervised form as test labels are unavailable. This work focuses on designing effective prompt formulations to provide the model with the context tailored to test samples, which helps precisely retrieve pretrained knowledge. As for $L_{tt}$, we directly utilize effective losses in existing TTA studies, which are thoroughly discussed in Section~\ref{sec::exp}.}

\textbf{Offline and online TTA.} Typical TTA scenarios involve offline and online setups. Offline TTA assumes that the entire target test distribution $\{x^t_i, i=1,..., N^t\}$ is available for training. \textcolor{black}{To prevent confusion, we distinguish between offline TTA and the traditional UDA setup: offline TTA is trained directly on a small test set, whereas UDA is trained on a larger and extra-collected training set of the target domain, together with labeled data from the source domain.}
For online TTA, test sample $x^t_i$ arrives sequentially at time step $i$. The target model $\mathcal{T}$ updates accordingly for the current sample ${\theta_{i-1}} \rightarrow {\theta_{i}}$ and makes the prediction on the fly $p^t_i = \mathcal{T}_{\theta_i}(x^t_i)$. \textcolor{black}{Coupled with proper parameter updating schemes, the proposed TTA framework is applicable to both online and offline situations.}

\textbf{Overview of the proposed method.} Despite the low-level style shift in $X^s$ and $X^t$, we first emphasize that the high-level shape discrepancy between $Y^s$ and unseen $Y^t$ also hampers the adaptation performance. In this regard, we devise two prompt modules: the input decorator (ID) and the cross-attention prompt modulator (CAPM), separately for reformulating the input data and learning target-specific shape information in the latent space, respectively. To obviate the overfitting of adaptation parameters to a particular test sample which might negatively affect the prediction of future samples in online TTA, we propose an alternating momentum updating (AMU) strategy. The overview of PASS is depicted in Figure \ref{fig::overview}, with details presented below.
% Figure 2 presents the overview of our PASS applied to the skip-connections of a U-shape encoder-decoder network. The key components are two prompts: theinput-space prompt to provide essential low-level information and the shape prompt to encode high-level degradation knowledge. 

\subsection{Input Decorator}
Previous studies have shown that leveraging prompts in the continual input embedding space introduces flexibility to pretrained models through an additional parameter space \cite{loedeman2022prompt, shu2022test}. However, current applications of visual prompt learning to domain adaptation \cite{hu2022prosfda, gan2023decorate} simply adopt the fixed prompt for all test samples which neglects the data variety within the target distribution. Hence, we propose to generate a dynamic prompt conditioned on each test sample. 
Let $\textrm{ID}$ be the data decorator parameterized by ${\phi_{D}}$. For each test data point $x^t_i$ at the $i$-th time step, $\textrm{ID}$ reformulates it as follows:
\begin{equation}
    \tilde{x}^t_i =  x^t_i + \textrm{ID}(x^t_i;\phi_{D}),
\end{equation}
where $\textrm{ID}$ intends to shift the distribution of target data $X^t$ to be close to the source data $X^s$, and $\tilde{x}^t_i$ refers to the altered target input. \textcolor{black}{Since Instance Normalization (IN) has shown effectiveness in style transfer\cite{huang2017arbitrary}, we adopt two shallow convolutional layers with IN to construct the ID.}

\subsection{Cross-Attention Prompt Modulator}
Introducing solely pixel-wise prompts to the pretrained model offers limited adaptation capacity for exploring convincing knowledge essential for semantic segmentation. Therefore, we 
further propose to leverage shape-aware information provided by the input image itself to boost the segmentation performance. In this regard, we construct shape prompts induced by high-level representations in the latent space.
%further propose to learn high-level prompts for each image to capture shape prior and the target-specific knowledge. 
Formally, given a reformulated test image $\tilde{x}^t_i$, let $z^t_i \in \mathbb{R}^{C\times H \times W}$ denote the latent features extracted by the encoder of $\mathcal{T}$ with
$C$ channels and size of $H \times W$. We set a learnable shape prompt bank $\overline{SP} \in \mathbb{R}^{L\times H \times W}$ parameterized by $\phi_{sp}$, consisting of $L$ shape prompt templates. 
$\overline{SP}$ is domain-shared for all test images within the target distribution. During the adaptation stage, the learnable shape prompt bank retrieves the knowledge of the pretrained model to obtain the target-relevant model prior.
Then, we propose a cross-attention prompt modulator $\textrm{CAPM}$ to generate the sample-specific shape prompt $SP_i$ from the enriched shape prompt bank $\overline{SP}$. 

Concretely, we regard the target feature $z^t_i$ as the dynamic query while the domain-common prompt bank $\overline{SP}$ as the key/value:
\begin{equation}
    Q_i = \phi_q(z^t_i),  K_i = \phi_k(\overline{SP}),  V_i = \phi_v(\overline{SP}).
\end{equation}
Here, we adopt lightweight depthwise convolutional layers for the mapping function $\phi_q, \phi_k,\phi_v$. Note that the mapping function keeps the output channels identical to the number of input. The dimension of $\{Q_i, K_i, V_i\}$ is flattened from $\mathbb{R}^{\frac{C}{L}\times H \times W}$ to $\mathbb{R}^{\frac{C}{L}\times HW}$.
Then we can perform a cross-attention operation between the target representation $z^t_i$ and the domain-common shape prompt bank $\overline{SP}$:
\begin{equation}
     \textrm{CAPM}(z^t_i, \overline{SP}) = \textrm{Softmax}((Q_i \times K_i^T) / \sqrt{L})V_i.
\end{equation}
Note that the attention map $A_i=\textrm{Softmax}((Q_i \times K_i^T)/ \sqrt{L})$ is in the transposed size of $\mathbb{R}^{C\times L}$ instead of the huge regular size $\mathbb{R}^{HW\times HW}$. Each row of $A_i$ indicates the attentive scores of a feature channel to $L$ shape prompt templates in $\overline{SP}$. To select the most related shape prompt templates to the current sample, we only reserve $top_k$ attention scores from the keys for more effective feature aggregation, leading to a sparse $A^*$:
\begin{equation}
    A_i^{*{p,q}} =
    \begin{cases}
        A_i^{p,q}  & \text{if }  A_i^{p,q} \geq top_k(A_i^p), \\
        0 & \text{Otherwise},
    \end{cases}
\end{equation}
where $p \in [1, C]$, $q \in [1, L]$. Finally, the modulated shape prompt is fused with the latent feature:
\begin{equation}
    \tilde{z}^t_i = {z}^t_i + SP_i =  {z}^t_i + \textrm{CAPM}(V_i, A^*_i;\phi_{M}),
\end{equation}
in which $\phi_{M} = \{\phi_{sp}, \phi_q, \phi_k, \phi_v\}$.  The cross-attention distillation leads to refined shape prompts and thus improves the feature generalization.

\subsection{Alternating Momentum Updating Strategy}
Online TTA is a challenging scenario where only one test sample is accessible at each adaptation step and the model is continually updated. If the current sample is far from the overall target distribution, the model updating from this sample is likely to cause detrimental effects on the remaining samples. To alleviate the instability caused by data variety and error accumulation, previous TTA approaches \cite{gan2023decorate, wang2022cotta} adopt the teacher-student network architecture for parameter updating. The student network $\mathcal{T}_{\theta_i}$ is online updated with the $i$-th sequentially arrived sample, whereas the weights of the teacher network $\mathcal{T}_{\theta'_{i-1}}\rightarrow \mathcal{T}_{\theta'_{i}}$ is updated by the exponential-moving-average (EMA) strategy:
\begin{equation}\label{eq::ema}
   \theta'_{i} = (1-m) {\theta'_{i-1}} + m  {\theta_{i}}.
\end{equation}

In this way, the teacher network accumulates new knowledge without forgetting historical knowledge. Usually, the updated parameters in the teacher model are used for predictions\cite{gan2023decorate, wang2022cotta}. However, we keep the teacher network as a buffer for historical knowledge restoration and use the student network for a quick adaptation to the test sample. At each step with new test data, the student network is reset by the teacher network and adapted accordingly to this data.
%We argue that the fixed momentum coefficient $m$ in EMA is suboptimal as the adaptation performance is either unstable or converges rather slowly. This
Moreover, we argue that the fixed momentum $m$ in EMA could cause the forgetting of source knowledge in long-term TTA. For stable adaptation and fast convergence, we propose to adapt the momentum with each incoming sample:\\
\begin{equation}\label{eq::m_decay}
   m_{i+1} = c + m_{i} \omega
\end{equation}
Where $c=0.005$ is a constant to ensure the lower bound of $m_{i}$ and $\omega \in [0, 1]$ is a decay factor. We set $m_0=0.1$ at the beginning of adaptation. As the momentum $m_i$ decays, the later samples will have a smaller impact, thereby avoiding the catastrophic forgetting problem. Our momentum decay strategy for updating trainable parameters actually shares a similar idea with the statistic updating scheme for BN layers in \cite{mirza2022dua}. The overall procedures of our PASS approach for online TTA are summarized in Algorithm \ref{algorithm}.

% \begin{algorithm}[htb]
% \caption{PASS in online test-time adaptation setting.}
% \label{algorithm}
% \begin{algorithmic}[1]
%     \Require
%     A source pretrained model $\mathcal{S}_{\psi^*}$, teacher/student models $\mathcal{T}_{\theta'}$/$\mathcal{T}_{\theta}$ equipped with ID/CAPM parameterized by $\phi$ ($\theta = \{\psi, \phi\}$), a test set $D^t=\{x^t_i\}$, hyperparameters $m$, $c$, the learning rate $\alpha$, the adaptation loss $L_{ada}$.
%     \State  $\theta'_{0}=\{\psi'_{0}, \phi'_0\} \gets \{\psi^*, \phi_{RandomInit}\}$\Comment{Initialize the teacher model with pretrained weights.}
%     \State  Set BN parameters and $\phi$ to be trainable and freeze other parameters in the student model $\mathcal{T}_{\theta}$.
%     \For{each time step $i$, current sample $x^t_i$}
%     \State $\theta_{i} \gets \theta'_{i-1}$ \Comment{Re-initialize student model with teacher weights.}
%     \State $\theta_{i} \gets \theta_{i} - \alpha \nabla L_{ada}(\mathcal{T}_{\theta_{i}}(x^t_i))$. \Comment{Update the student model via the unsupervised adaptation loss.}
%     \State Update $\theta'_{i}$ by Equation \ref{eq::ema} with $m_i$.\Comment{Update the teacher model.}
%     \State Adapt the momentum $m_{i+1}$ by Equation \ref{eq::m_decay}. \Comment{Decrease the momentum factor.}
%     \State $p^t_i \gets \mathcal{T}_{\theta_{i}}(x^t_i)$ \Comment{Get test-time predictions via the adapted student model.}
%     \EndFor
%     \State \textbf{Return} all student model predictions and teacher model parameters.
% \end{algorithmic}
% \end{algorithm}

\begin{algorithm}[!t]
\caption{PASS in online test-time adaptation setting.}
\label{algorithm}
% \begin{algorithmic}[1]
    \textbf{Initialize:}
    A teacher model $\mathcal{T}_{\theta'_0}$ ($\theta'_0 = \{\psi'_0, \phi'_0\}$) where the backbone parameters $\psi'_0$ are initialized from pretrained weights $\psi^*$ and the adaptation parameters $\phi'_0$ are randomly initialized, hyperparameters $m$ and $\omega$, the learning rate $\alpha$ and the test-time training loss $L_{tt}$;\\
    \textbf{Input:} For each time step $i$, current test sample $x^t_i$.
    \begin{algorithmic}[1]
\State $\theta_{i} \gets \theta'_{i-1}$. \Comment{Reset the student model with teacher weights.}
\State  Set $\phi_i$ and BN parameters to be trainable in the student model $\mathcal{T}_{\theta_i}$ with others frozen.
\State $\theta_{i} \gets \theta_{i} - \alpha \nabla L_{tt}(\mathcal{T}_{\theta_{i}}(x^t_i))$. \Comment{Update the student model via the unsupervised adaptation loss.}
\State $p^t_i \gets \mathcal{T}_{\theta_{i}}(x^t_i)$. \Comment{Get test-time predictions via the adapted student model.}
\State Update $\theta'_{i}$ by Equation \eqref{eq::ema} with $m_i$.\Comment{Update the teacher model via moving average.}
\State Adapt the momentum $m_{i+1}$ by Equation \eqref{eq::m_decay}. \Comment{Decrease the momentum factor.}
\end{algorithmic}
    \textbf{Output:} All predictions and teacher model parameters.
% \end{algorithmic}
\end{algorithm}

\section{Experiments}
\label{sec::exp}
%In this section, we first introduce the datasets and implementation details. Then, we present detailed comparisons and analyses.
\subsection{Experimental Setup}
%\textbf{Datasets and Metrics.}
\textbf{The Fundus OD/OC Segmentation.} 
We first evaluate our method on the joint optic disc (OD) and optic cup (OC) segmentation task. We use a multi-domain RIGA+ dataset \cite{almazroa2018retinal, decenciere2014feedback, hu2022prosfda} comprising five domains derived from different resources: BinRushed, Magrabia, BASE1, BASE2, and BASE3. 
The 195 and 95 labeled images in BinRushed and Magrabia serve as the source domain while 35, 30 and 27 unlabeled images in three test sets of BASE1, BASE2, and BASE3 are utilized as target domains, respectively. Following \cite{hu2022prosfda}, each image was resized to 512$\times$512 pixels and the intensities are normalized to a distribution with zero mean and unit variance. 

\textbf{The MRI Prostate Segmentation.} We also validate our method on the prostate segmentation task with T2-weighted MRI scans collected from six clinical centers \cite{nicholas2015nci, litjens2014evaluation, lemaitre2015computer}: RUNMC, BMC, I2CVB, UCL, BIDMC and HK (Sites A-F). We use 30 labelled cases in RUNMC as the source dataset and 30/19/13/12/12 unlabeled cases in the remaining five sites for target testing. Each axial slice of the MRI volume was resized to 384$\times$384 pixels and normalized to zero mean and unit variance in intensity values. Due to the large variance in the slice thickness of data, we employ a 2D network as the backbone. For all MRI slices, we clipped 5\%-95\% of the histograms before the intensity normalization.

\textbf{Implementation Details.}
All the experiments are run with the PyTorch framework on one NVIDIA-V100 GPU. We adopt a 2D U-Net with ResNet-34\cite{ronneberger2015u} as the backbone for OD/OC and prostate datasets, following previous TTA works\cite{hu2022prosfda, chen2023vptta}.
In the pretraining stage, the models are trained on the source data using the binary cross entropy (BCE) and dice losses with a batch size of 16. We employ data augmentation schemes of elastic deformation, random rotation, scaling, flipping, Gaussian noises, brightness multiplicative, Gamma and contrast transformations.
We use the Adam optimizer with an initial learning rate of 0.001/0.0001 for 100/800 epochs on OD/OC and prostate datasets, respectively. Empirically, we chose the model trained at the last epoch as the testing model.
\textcolor{black}{During the optimization stage, we select the BN loss \cite{hu2022prosfda} of aligning the source-target statistics and the TENT loss \cite{wang2021tent} of minimizing the output entropy.}
%We selected the BN loss \cite{hu2022prosfda} of aligning the source-target statistics for OD/OC segmentation tasks and the TENT loss \cite{wang2021tent} for prostate segmentation tasks following \cite{yang2022dltta, bateson2022source}. The detailed comparison of different losses is discussed in the ablation study. 
Note that only the BN parameters and the introduced adaptation parameters were trainable and the rest of the parameters were frozen after being initialized by the pretrained model. We compare our PASS with previous approaches on the offline and online TTA setups separately. Under the offline TTA setting, we train the target model on the entire test dataset with a batch size of 8/16 for 20/100 epochs on OD/OC and prostate datasets, respectively. Under the online TTA setting, we perform continual adaption for each single test subject. The Adam optimizer was utilized with a learning rate of 0.005 and 0.01 for the OD/OC and prostate segmentation tasks, respectively. The hyperparameters $L$ (size of the shape prompt bank), $k$ ($top_k$ for the sparsity of $A$), and $\omega$ (decay coefficient in AMU) are set to 512, 0.1 and 0.94/0.6 for OD/OC and prostate segmentation tasks. \changed{M1.2}{\link{R1.2}}{The average response time of PASS on the OD/OC and prostate segmentation tasks was 0.36s and 1.81s, respectively.}

\begin{table}[!t]
\renewcommand{\arraystretch}{0.8} 
\setlength{\extrarowheight}{2pt}
\setlength{\tabcolsep}{4pt} % 调整列间距
\centering
\caption{Offline TTA performance of different methods in DSC Sore (\%) on three OD/OC segmentation test sets.}
% \resizebox{1\textwidth}{!}{
\begin{tabular}{lllllllc}
\toprule
\multirow{2}{*}{Method} & \multicolumn{2}{c}{BASE1}                & \multicolumn{2}{c}{BASE2}                 & \multicolumn{2}{c}{BASE3}                         & \multirow{2}{*}{Average} \\ \cmidrule(r){2-3}\cmidrule(r){4-5}\cmidrule(r){6-7}
                        & \multicolumn{1}{c}{OD}                  & \multicolumn{1}{c}{OC}                  & \multicolumn{1}{c}{OD}                  & \multicolumn{1}{c}{OC}                  & \multicolumn{1}{c}{OD}                  & \multicolumn{1}{c}{OC}                  &                 \\ \midrule
Source Only        & 95.21          & {\ul 85.20}    & 92.58         & 77.73         & 93.72          & 82.16         & 87.77\scriptsize{±6.49}                 \\ \midrule
TENT\cite{wang2021tent}                  & 95.34          & 82.30          & 94.80          & 78.30          & 94.15        & 81.39          & 87.71\scriptsize{±7.16}                 \\
RN-CR\cite{hu2021rncr}                  & 95.15         & 80.55          & 94.16          & 76.39         & {\ul 94.24}         & 80.80          & 86.88\scriptsize{±7.77}               \\
AdaMI\cite{bateson2022moment}                  & \textbf{95.42}    & 84.94         & {\ul 94.89}   & 79.21          & {94.19} & 82.84          & 88.58\scriptsize{±6.48}                       \\
OCL\cite{zhang2023oclttt}                    & 94.61         & 81.80          & 94.73          & {\ul 83.07}  & 94.04         & {\ul 84.51}    & 88.79\scriptsize{±5.72}                 \\
ProSFDA\cite{hu2022prosfda}                & {\ul 95.37} & \textbf{85.40} & 94.58          & 81.43         & 93.91          & 82.92         & {\ul 88.94\scriptsize{±5.82}}                 \\ \midrule
PASS (ours)            & 95.18          & 84.52          & \textbf{95.25} & \textbf{83.99} & \textbf{94.75}    & \textbf{85.52} & \textbf{89.87\scriptsize{±5.21}}        \\ \bottomrule
\end{tabular}
\label{tab::Offline_TTA_RIGA}
\end{table}
\begin{figure}[t]
\centering
\includegraphics[width=0.95\linewidth]{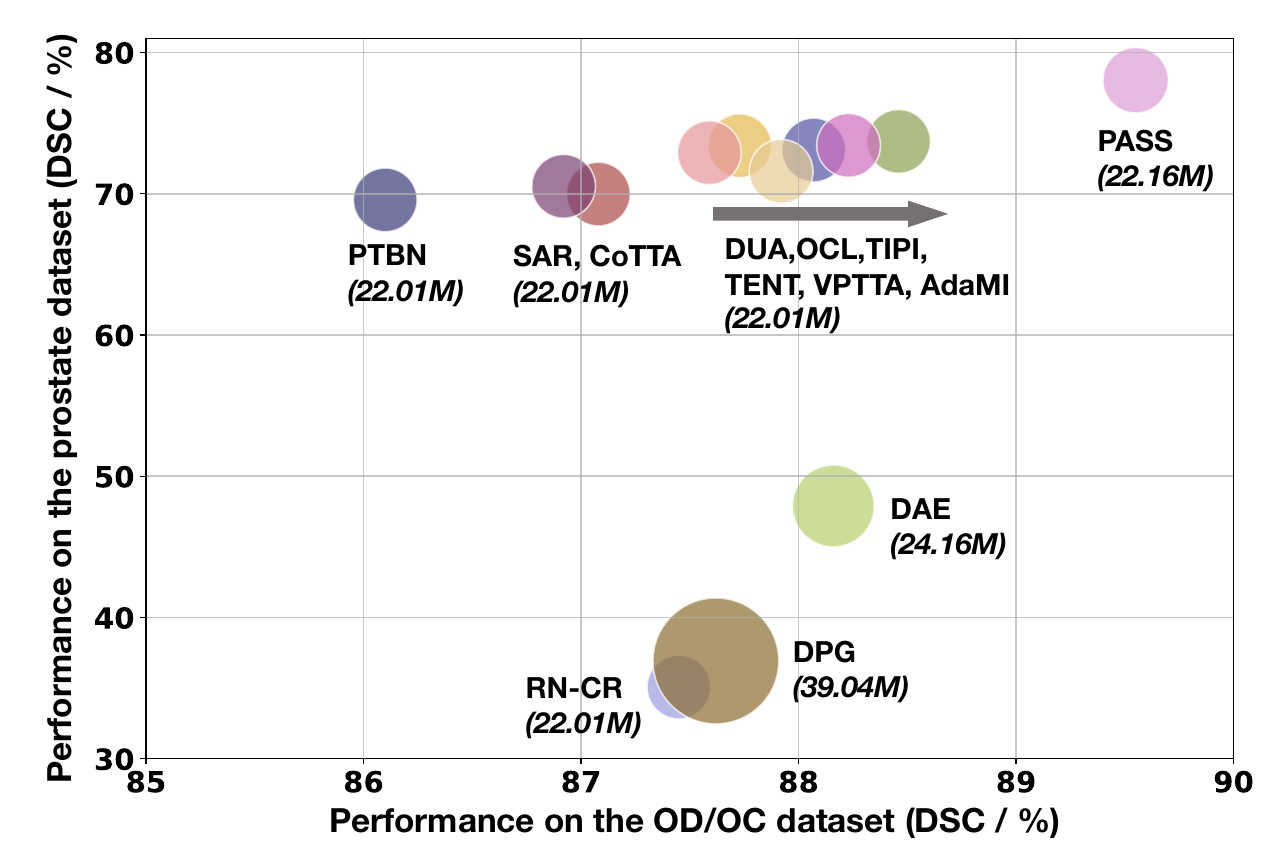}
\caption{Graphical comparison of the online TTA performance and the number of parameters (M) among different TTA models on the OD/OC and prostate datasets.}
\label{fig::params}
\end{figure}
\begin{table}[t!]
\renewcommand{\arraystretch}{0.8} 
\setlength{\extrarowheight}{2pt}
\centering
\caption{Online TTA performance of different methods in DSC Score (\%) on three OD/OC segmentation test sets.}
\setlength{\tabcolsep}{4pt} % 调整列间距
%>{\columncolor{gray!20}}
% \resizebox{1\textwidth}{!}
\begin{tabular}{l@{\hspace{0.6em}}ccccccc}
\toprule
\multirow{2}{*}{Method} & \multicolumn{2}{c}{BASE1}                & \multicolumn{2}{c}{BASE2}                 & \multicolumn{2}{c}{BASE3}                         & \multirow{2}{*}{Average} \\ \cmidrule(r){2-3}\cmidrule(r){4-5}\cmidrule(r){6-7}
                        & \multicolumn{1}{c}{OD}                  & \multicolumn{1}{c}{OC}                  & \multicolumn{1}{c}{OD}                  & \multicolumn{1}{c}{OC}                  & \multicolumn{1}{c}{OD}                  & \multicolumn{1}{c}{OC}                  &                    \\ \midrule
Source Only         & 95.21          & {\ul 85.20}    & 92.58         & 77.73         & 93.72          & 82.16         & 87.77\scriptsize{±6.49}                  \\ 
 \midrule
PTBN\cite{nado2020ptbn}                  & 94.21          & 80.87          & {\ul 94.51}        & 74.95          & 93.05        & 79.02          & 86.10\scriptsize{±8.03}                \\
TENT\cite{wang2021tent}                  & 95.05          & 84.19          & 93.42          & 79.07          & 93.66        & 83.05          & 88.07\scriptsize{±6.19}                  \\
RN-CR\cite{hu2021rncr}                  & 95.27         & 84.67          & 94.35          & 75.70         & 94.11          & 80.57          &87.45\scriptsize{±7.60}              \\
AdaMI\cite{bateson2022moment}                  & {95.05}    & {85.07}        & {93.62}   & 79.61          & {93.97} & 83.43          &{\ul88.46\scriptsize{±5.99}}   \\
DAE\cite{karani2021dae}           & 93.82         & 82.71          & 92.23          & {\ul 81.24}  & 94.08         & \textbf{84.86}    & 88.16\scriptsize{±5.36}                  \\
DPG\cite{valanarasu2022dpg}           & 95.26         & 84.14          & 93.45          & {78.11}  & 93.76         & {80.97}    & 87.62\scriptsize{±6.79}                \\
OCL\cite{zhang2023oclttt}                    & 95.11         & 84.08          & 94.11          & {77.58}  & 94.07         & {81.41}    & 87.73\scriptsize{±6.97}               \\
TIPI\cite{yang2022dltta}                & {\ul 95.43} & {84.67} & 93.22          & 77.60          & 94.05          & 82.56         & 87.92\scriptsize{±6.68} 
  \\
CoTTA\cite{wang2022cotta}                & \textbf{95.62} & {82.98} & 94.36          & 75.12          & 93.93          & 80.47         & 87.08\scriptsize{±7.92} 
  \\
DUA\cite{mirza2022dua}                & {95.39} & {84.06} & 93.53          & 74.97          & {\ul 94.43}          & 83.21         & 87.59\scriptsize{±7.46} 
  \\
SAR\cite{niu2023sar}                & {95.24} & {82.37} & {94.37}        & 75.11          & 93.93          & 80.48         & 86.92\scriptsize{±7.91} 
\\ 
VPTTA\cite{chen2023vptta}                & {95.22} & {84.45} & {94.30}        & 78.12          & 94.42          & 82.88         & 88.23\scriptsize{±6.70} 
\\\midrule
PASS (ours)            & 95.26          & \textbf{85.33}          & \textbf{95.12} & \textbf{83.18} & \textbf{94.61}    & {\ul 83.77} & \textbf{89.55\scriptsize{±5.49}}        \\ \bottomrule
\end{tabular}
\label{tab::Online_TTA_RIGA}
\end{table}
\begin{table*}[t]
\renewcommand{\arraystretch}{0.8} 
\centering
\caption{Offline TTA performance of different methods in DSC Sore (\%) and the 95th percentile of the Hausdorff Distance (HD95) on five prostate segmentation test sets.}
\setlength{\extrarowheight}{3pt}
% \setlength{\tabcolsep}{1pt} % 调整列间距
% \resizebox{1\textwidth}{!}{
% \adjustbox{max width=\textwidth}{
\begin{tabular} {l@{\hspace{0.6em}}llllllllllcl}
% \begin{tabularx}{\textwidth}
% {X|@{\hspace{0.6em}}XXXXXXXXXX|XX}
\toprule
\multirow{2}{*}{Method} & \multicolumn{2}{c}{B (BMC)}                              & \multicolumn{2}{c}{C (I2CVB)}                              & \multicolumn{2}{c}{D (UCL)}                              & \multicolumn{2}{c}{E (BIDMC)}                 & \multicolumn{2}{c}{F (HK)}              & \multicolumn{2}{c}{Average}             \\ \cmidrule(r){2-3}\cmidrule(r){4-5}\cmidrule(r){6-7}\cmidrule(r){8-9}\cmidrule(r){10-11}\cmidrule(r){12-13}%\cline{2-13} 
                        & \multicolumn{1}{c}{DSC$\uparrow$} & \multicolumn{1}{c}{HD95$\downarrow$} & \multicolumn{1}{c}{DSC$\uparrow$} & \multicolumn{1}{c}{HD95$\downarrow$} & \multicolumn{1}{c}{DSC$\uparrow$} & \multicolumn{1}{c}{HD95$\downarrow$} & \multicolumn{1}{c}{DSC$\uparrow$} & HD95$\downarrow$        & \multicolumn{1}{c}{DSC$\uparrow$} & HD95$\downarrow$      & DSC$\uparrow$       & \multicolumn{1}{c}{HD95$\downarrow$} \\ \midrule
Source Only          & 78.24              & 14.14              & 69.76             & 31.09              & 80.41              & {\ul 14.68}             & 48.49             & 54.03 & 84.90              & 3.13 & 72.36\scriptsize{±12.91}     &23.41\scriptsize{±17.72}           \\ \midrule
TENT\cite{wang2021tent}                    &  82.68                       &   {\ul 4.78}                        &    {72.34}                    &  24.43                       &     80.51                      & 17.43               &  55.04                     &   62.09          &   \textbf{86.48}                &  \textbf{2.66}          &   75.41\scriptsize{±11.19}         &  {22.28\scriptsize{±21.47} }                 \\
RN-CR\cite{hu2021rncr}                    &   57.35                      &    57.17                       &     32.79                 &      141.03                     &     64.95                     &       38.27                    &     48.73                    &       61.17       &  85.17                       &   4.22          &   57.80\scriptsize{±17.37}         &  60.37\scriptsize{±45.07}                 \\
AdaMI\cite{bateson2022moment}                   & {\ul 82.80}                   & 6.08                          & {\ul 72.74}                  &  {\ul 24.36}                   & 80.90                &  {15.62}                         &  {\ul 59.04}                        &   47.41           &  86.36                       &  2.69          &  {\ul 76.37\scriptsize{±9.75} }        &  {\ul 19.23\scriptsize{±16.00} }                   \\
OCL\cite{zhang2023oclttt}                     &78.46                          & 12.92                          &71.55                          & 29.10                    & \textbf{81.65}                         & \textbf{12.84}                          &  51.83                        &  52.75            & 84.96                  &3.02           &73.69\scriptsize{±11.79}             &  22.13\scriptsize{±17.45}                   \\
ProSFDA\cite{hu2022prosfda}                 & 74.60                      & 14.95                       & 53.66                         & 28.23                          & 77.50                         & 21.23                          & 48.35                         & {\ul 43.77}             &  85.50                     &  2.91          &    67.92\scriptsize{±16.06}         &  22.22\scriptsize{±13.62}                  \\ \midrule
PASS (ours)            & \textbf{83.28}                         &  \textbf{4.49}                         & \textbf{75.36}              & \textbf{20.11}              & {\ul 80.92}                  & 18.29                          & \textbf{72.43}                         &\textbf{25.74}              & {\ul 86.42}                & {\ul 2.67}           & \textbf{79.68\scriptsize{±5.12} }  &   \textbf{14.26\scriptsize{±9.08} }          \\ \bottomrule
\end{tabular}
\label{tab::Offline_TTA_Prostate}
\end{table*}
\begin{table*}[t]
\renewcommand{\arraystretch}{0.8} 
\centering
\caption{Online TTA performance of different methods in DSC Sore (\%) and the 95th percentile of the Hausdorff Distance (HD95) on five prostate segmentation test sets.}
\setlength{\extrarowheight}{2pt}
% \setlength{\tabcolsep}{2pt} % 调整列间距
% \resizebox{1\textwidth}{!}{
% \scalebox{0.52}{
% @{\hspace{0.6em}}
\begin{tabular}{lllllllllllcl}
\toprule
\multirow{2}{*}{Method}& \multicolumn{2}{c}{B (BMC)}                              & \multicolumn{2}{c}{C (I2CVB)}                              & \multicolumn{2}{c}{D (UCL)}                              & \multicolumn{2}{c}{E (BIDMC)}                 & \multicolumn{2}{c}{F (HK)}    & \multicolumn{2}{c}{Average}             \\ \cmidrule(r){2-3}\cmidrule(r){4-5}\cmidrule(r){6-7}\cmidrule(r){8-9}\cmidrule(r){10-11}\cmidrule(r){12-13}%\cline{2-13} 
                        & \multicolumn{1}{c}{DSC$\uparrow$} & \multicolumn{1}{c}{HD95$\downarrow$} & \multicolumn{1}{c}{DSC$\uparrow$} & \multicolumn{1}{c}{HD95$\downarrow$} & \multicolumn{1}{c}{DSC$\uparrow$} & \multicolumn{1}{c}{HD95$\downarrow$} & \multicolumn{1}{c}{DSC$\uparrow$} & HD95$\downarrow$        & \multicolumn{1}{c}{DSC$\uparrow$} & HD95$\downarrow$      & DSC$\uparrow$       & \multicolumn{1}{c}{HD95$\downarrow$} \\ \midrule
Source Only        & 78.24              & 14.14              & 69.76             & {31.09}              & 80.41              & {14.68}             & 48.49             & 54.03 & 84.90              & 3.13 & 72.36\scriptsize{±12.91}    & 23.41\scriptsize{±17.72}         \\ \midrule
PTBN\cite{nado2020ptbn}                    &  78.19                       &   {17.90}                        &    {65.56}                    &  47.36                       &     77.11                      & 21.95               &  42.30                     &   58.34          &   84.71                &  3.37          &   69.57\scriptsize{±14.97}
        &     29.78\scriptsize{±20.12}                     \\
TENT\cite{wang2021tent}                    &  78.54                       &   {11.64}                        &    {69.71}                    &  31.38                       &     {80.64}                      & {14.55}               &  51.72                     &   {\ul 27.01}        &   84.91                &  3.09          &   73.10\scriptsize{±11.79}       &   {\ul 17.53\scriptsize{±10.33}}                      \\
RN-CR\cite{hu2021rncr}                    &   21.17                      &    134.92                       &     17.02                 &      165.59                     &     37.59                     &       95.63                    &     32.07                    &       101.78       &  67.57                        &   96.67          &   35.08\scriptsize{±17.84}        &   118.92\scriptsize{±27.44}                    \\
AdaMI\cite{bateson2022moment}                   & {78.46}                   & 16.23                          & {68.78}                  &  {39.93}                   & 80.81                &  {13.00}                        &  {\ul55.44}                       &   44.67           &  {\ul 85.04}                      &  3.18          &  {\ul 73.71\scriptsize{±10.58}}         &     {23.40\scriptsize{±16.09}}                     \\
DAE\cite{karani2021dae}                   & {57.96}                   & 13.82                          & {31.26}                  &  {105.60}                   & 63.86                &  {\ul 11.17}                        &  {31.77}                       &   72.87           &  54.63                        &  15.13          &  {47.90\scriptsize{±13.70}}         &     {43.72\scriptsize{±38.60}}                     \\
DPG\cite{valanarasu2022dpg}                   & {48.13}                   & 143.19                          & {29.91}                  &  {101.67}                   & 43.31                &  {94.38}                        &  {17.54}                       &   141.66           &  45.77                        &  124.53          &  {36.93\scriptsize{±11.57}}         &     {121.09\scriptsize{±20.06}}                     \\
OCL\cite{zhang2023oclttt}                     &75.77                          & 15.29                          &69.27                          & 31.56                    & \textbf{81.79}                         & {12.57}                          &  55.25                        &  41.41            & 84.91                  &{\ul 2.99}           &73.40\scriptsize{±10.54}            &  20.76\scriptsize{±13.83}                  \\
TIPI\cite{nguyen2023tipi}                 & 79.06               & {10.30}               & 68.96                         & {\ul 30.68}                         & 80.56                         & \textbf{9.89}                          & 44.61                         & {52.91}             &  84.76                       &  3.26          &    71.59\scriptsize{±14.45}
       &          21.41\scriptsize{±18.24}                \\ 
       CoTTA\cite{wang2022cotta}                 & 78.37               & 16.32               & 66.52                         & 46.44                          & 77.15                         & 21.46                          & 43.24                         & {56.46}             &  84.63                       &  3.32          &   69.98\scriptsize{±14.58}
       &           28.80\scriptsize{±19.67}       \\
        DUA\cite{mirza2022dua}                 & {\ul 79.20}              & {\ul 9.86}               & {\ul 69.80}                         & 33.19                          & 79.91                      & 11.91                     & 50.89                         & {56.73}             &  84.75                    &  3.34       &    72.91\scriptsize{±12.03}
       &            23.01\scriptsize{±19.61}             \\
       SAR\cite{niu2023sar}                 & 76.42               & 24.76               & 63.88                      & 48.76                         & 79.09                        & 18.93                         & 48.49                         & {54.03}             &  84.75                       &  3.34          &    70.53\scriptsize{±12.96}
       &          29.96\scriptsize{±18.92}               \\ VPTTA\cite{chen2023vptta}  
       
       & 78.64              & 12.52              & 70.93                      & 33.33                         & 80.52                         & 14.39                         & 52.14                         & {53.72}             &  {84.94}                       &  {3.12}          &    73.43\scriptsize{±11.57}
       &          23.42\scriptsize{±18.05}        \\
       
       \midrule
PASS (ours)            & \textbf{79.26}                         &  \textbf{9.80}                         & \textbf{72.64}              & \textbf{28.79}              & {\ul 81.36}                  & 14.99                         & \textbf{71.68}                         &\textbf{21.05}              & \textbf{85.27}               & \textbf{2.72}           & \textbf{78.04\scriptsize{±5.18}} & \textbf{15.47\scriptsize{±8.98}}            \\ \bottomrule
\end{tabular}
\label{tab::Online_TTA_Prostate}
\end{table*}

\subsection{Performance Comparison}
\textbf{Comparison methods.} We compare our PASS with the ``Source Only'' baseline (directly applying the model trained on the source domain to the target domain), and thirteen TTA methods, including two BN variants (PTBN\cite{nado2020ptbn} and DUA\cite{mirza2022dua}), three entropy-minimization methods (TENT\cite{wang2021tent}, RN-CR\cite{hu2021rncr} and SAR\cite{niu2023sar}), a contrastive learning-based method
(OCL\cite{zhang2023oclttt}), a pseudo-label-based method (CoTTA\cite{wang2022cotta}), a method using the transform invariance regularizer as the surrogate loss (TIPI\cite{nguyen2023tipi}), two visual prompting methods with the BN statistics alignment loss (ProSFDA\cite{hu2022prosfda} and VPTTA\cite{chen2023vptta}), a source-prior-guided method (AdaMI\cite{bateson2022moment}) based on entropy minimization
%, a method dynamically adjusting the learning rate (DLTTA\cite{yang2022dltta}),
and two auto-encoder-based methods (DAE\cite{karani2021dae} and DPG\cite{valanarasu2022dpg}). Depending on the applicable situations of different methods, we choose twelve for online TTA and five for offline TTA. \changed{M3.2}{\link{R3.2}}{All the comparison methods are based on the same baseline as our PASS, with some introducing additional modules. We first show a graphical comparison
in terms of online TTA performance and model size in Figure \ref{fig::params}. Then, we provide detailed comparisons for offline TTA and online TTA.}

\begin{figure*}[t]
\centering
\includegraphics[width=1\linewidth]{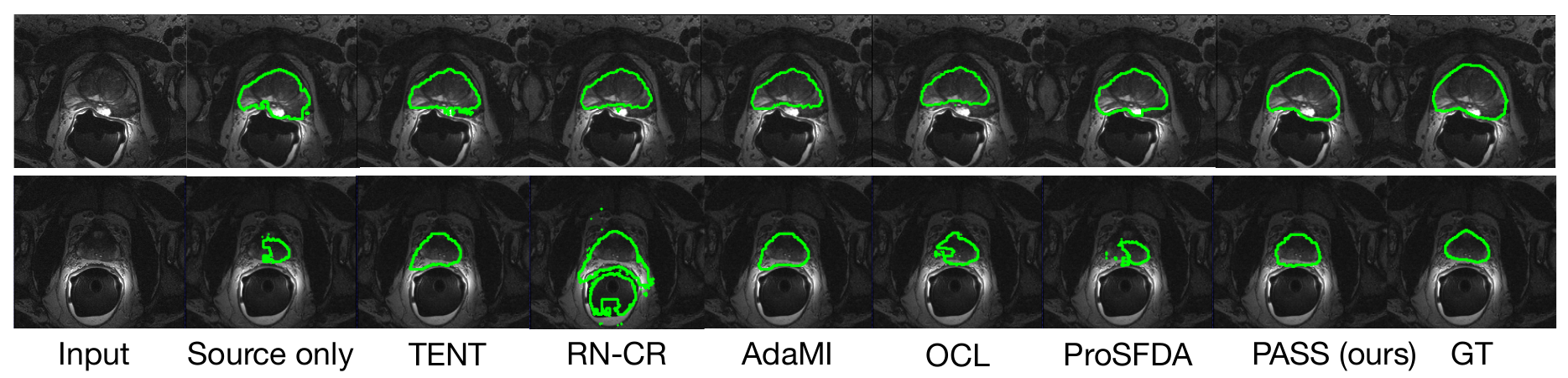}
\caption{Visualization of segmentation results generated by different approaches in prostate test-time adaptation applications.}
\label{fig::prostate_seg}
\end{figure*}

\textbf{Results on the OD/OC segmentation task.} 
First, we conducted the comparison experiments on the joint OD/OC segmentation adaptation task from a source dataset to three target sets BASE1-3. As claimed in \cite{hu2022prosfda}, the domain discrepancy between the source dataset and BASE1 is relatively smaller than BASE2/3 according to the performance gap between ``Intra-domain'' (vanilla training and testing within the same domain) to ``Source Only'' baseline. For BASE1, ``Source Only'' has reached a satisfactory level due to the narrow domain shift. Hence, effective TTA methods mainly enhance the performance of BASE2/3 while maintaining the performance of BASE1.
The results in the offline TTA setting in Table \ref{tab::Offline_TTA_RIGA} show that PASS achieves consistently better performance than other competing methods on BASE2/3. The entropy-based TENT and RN-CR fail to improve the baseline on the average dice while the AdaMI, OCL and ProSFDA yield performance gains. This indicates that entropy minimization is a weak supervision and it is imperative to introduce regularization terms that consider the informative knowledge in semantic segmentation. Compared to ProSFDA which only learns the same visual prompt for all samples, the proposed PASS involves a dynamic sample-dependent visual prompt and a high-level shape prompt, leading to the best average performance across all target domains. Table \ref{tab::Online_TTA_RIGA} presents the results of the online adaptation setting. For this more challenging scenario, the majority of approaches only perform well on the optic disc (OD) class but yield frustrating results on the optic cup (OC) class. The OD is easier to segment since its boundary is much clearer than the OC.
AdaMI and DAE mitigate this issue by introducing a shape prior loss and pseudo anatomical labels generated by pretrained denoising auto-encoder from the source domain. Using shape-relevant information can facilitate the adaptation stage but the utility of these methods is greatly limited by the need for source data re-access or dedicated pretraining modules. Our PASS directly distills high-order shape prompts from the target data itself and achieves remarkable performance over other baselines.
\begin{table}[t]
\centering
\caption{The effects of different components of PASS on the OD/OC and prostate segmentation tasks in both online and offline TTA settings. The reported value is the average Dice Sore (DSC \%) on three OD/OC test sets or five prostate test sets (mean±std). When the AMU is excluded (w/o AMU), an independent adaptation is performed for each test sample.}
\renewcommand{\arraystretch}{1} 
\setlength{\tabcolsep}{8pt} % 调整列间距
%\resizebox{0.9\textwidth}{!}{
% \begin{tabular}{l@{\hspace{0.6em}}lll@{\hspace{0.9em}}lll}
\begin{tabular}{llll}
\toprule
\multicolumn{1}{l}{\multirow{2}{*}{Method}} & \multicolumn{3}{c}{Online TTA Setup}                                                                                    \\ %\cline{2-9}
\cmidrule(r){2-3}\cmidrule(r){4-4}
\multicolumn{1}{c}{}                        & \multicolumn{1}{c}{OD} & \multicolumn{1}{c}{OC}  & \multicolumn{1}{c}{Prostate} \\ \midrule
PASS                                       & \textbf{95.00\scriptsize{±0.28}}                    & \textbf{84.09\scriptsize{±0.91}}                                        & \textbf{78.04\scriptsize{±5.18}}                                      \\ \midrule
w/o ID                                       & 94.43\scriptsize{±0.48}                     & 81.51\scriptsize{±1.61}                     &  72.66\scriptsize{±13.90}                                                  \\ 
w/o CAPM                                   

 & 94.32\scriptsize{±0.65}                     & 82.87\scriptsize{±1.93}                                         & 74.34\scriptsize{±10.45} 
 \\ \midrule
{w/o AMU}                             

 & 94.46\scriptsize{±0.40}                     & 78.70\scriptsize{±6.03}                                         & 74.42\scriptsize{±6.08} 
 \\
 w/ EMA                                  

 & 94.18\scriptsize{±0.75}                     & 82.64\scriptsize{±1.22}                                         & 72.40\scriptsize{±13.77}
 \\\midrule
\multicolumn{1}{l}{\multirow{2}{*}{Method}} & \multicolumn{3}{c}{Offline TTA Setup}                                                                                            \\ %\cline{2-9}
\cmidrule(r){2-3}\cmidrule(r){4-4}
\multicolumn{1}{c}{}                        & \multicolumn{1}{c}{OD} & \multicolumn{1}{c}{OC}  & \multicolumn{1}{c}{Prostate} \\ \midrule
PASS                                       &     \multicolumn{1}{l}{\textbf{95.06\scriptsize{±0.22}}}             & \textbf{84.68\scriptsize{±0.63}}             & \textbf{79.68\scriptsize{±5.12}}                   \\ \midrule
w/o ID              &  \multicolumn{1}{l}{94.94\scriptsize{±0.48}}             & 80.17\scriptsize{±3.41}             & 76.56\scriptsize{±9.81}                   \\ 
w/o CAPM                                   
& \multicolumn{1}{l}{95.03\scriptsize{±0.13}}             & 81.67\scriptsize{±2.97}             & 78.38\scriptsize{±6.35} \\ \bottomrule
% w/o AMU                                      & 94.46\scriptsize{±0.40}             & 74.70\scriptsize{±6.04}             & 66.62\scriptsize{±14.65}                   & \multicolumn{3}{c}{—}                                                                                       \\ \bottomrule
\end{tabular}
\label{tab::ablation}
\end{table}

\textbf{Results on the prostate segmentation task.} In Table \ref{tab::Offline_TTA_Prostate}, we compare PASS with the existing methods on the multi-center MRI prostate segmentation adaptation task from Site A to Sites B-F in an offline test-time adaptation setting. Benefiting from the abundant information provided by diverse slices in the entire test set, significant performance improvement is obtained by most of the TTA algorithms, except for RN-CR and ProSFDA. The Contour Regularization (CR) in RN-CR is an unsupervised loss punishing the prediction difference of pixels and their neighbours within a radius. This is inappropriate for prostates with non-circular shapes and thus results in model collapse. Despite the excellent behaviours of ProSFDA on the OD/OC segmentation tasks shown in Table \ref{tab::Offline_TTA_RIGA}, it fails in prostate segmentation with more variable shapes. 
%Both the image style shift and the shape inconsistency (see Figure \ref{fig::shape_descriptor}) in prostate datasets are larger than in fundus datasets. For slight domain shifts, aligning the BN statistics is a reasonable strategy for better transfer performance. However, when the domain shift is substantial, forcibly imposing statistic alignment could hinder the adaptation stage. 
Our PASS significantly exceeds other methods on the overall performance across five domains, underscoring its effectiveness and robustness. 
Under the online TTA setting in Table \ref{tab::Online_TTA_Prostate}, only TENT, AdaMI, OCL, DUA, VPTTA and PASS outperform the “Source Only” baseline. This suggests the difficulty of adaptation towards the online distribution shifts when only one case is accessible at a time. Compared to the offline TTA, RN-CR causes more severe model crashes as the circular constraint is violated due to the diverse and irregular shapes of prostates in 2D slices. Besides, DAE and DPG that rely on pretrained auto-encoders in the source domain as the domain-common priors yield poor results. In scenarios with a relatively small domain gap (e.g., OD/OC datasets), pretrained auto-encoders can deliver some meaningful fine-grained knowledge that is shareable with the target data, but in large-gap situations, they may cause model instability or collapse. We observe that TENT/AdaMI only provide obvious performance increases in Site E (BIDMC) 
%(1) the target domain that possesses the most similar shape descriptors with the source domain as illustrated in Figure \ref{fig::shape_descriptor}, i.e. Site D (UCL) and F (HK) (2) 
which possesses a low shape similarity with the source domain (see Figure \ref{fig::shape_descriptor}). The transfer performance can be particularly poor without any constraints. Thus, even simple entropy minimization and inaccurate shape prior losses can boost the transferability. Among all the competing methods, our PASS stands out as the only one that improves performance substantially across all domains, proving its superiority against
varying target distribution shifts. Figure \ref{fig::prostate_seg} shows the segmentation results and proves that PASS can precisely preserve the anatomical shapes at a new target distribution, whereas other methods sometimes fail to do so.
\begin{figure*}[t]
    \centering
    \includegraphics[width=1\linewidth]{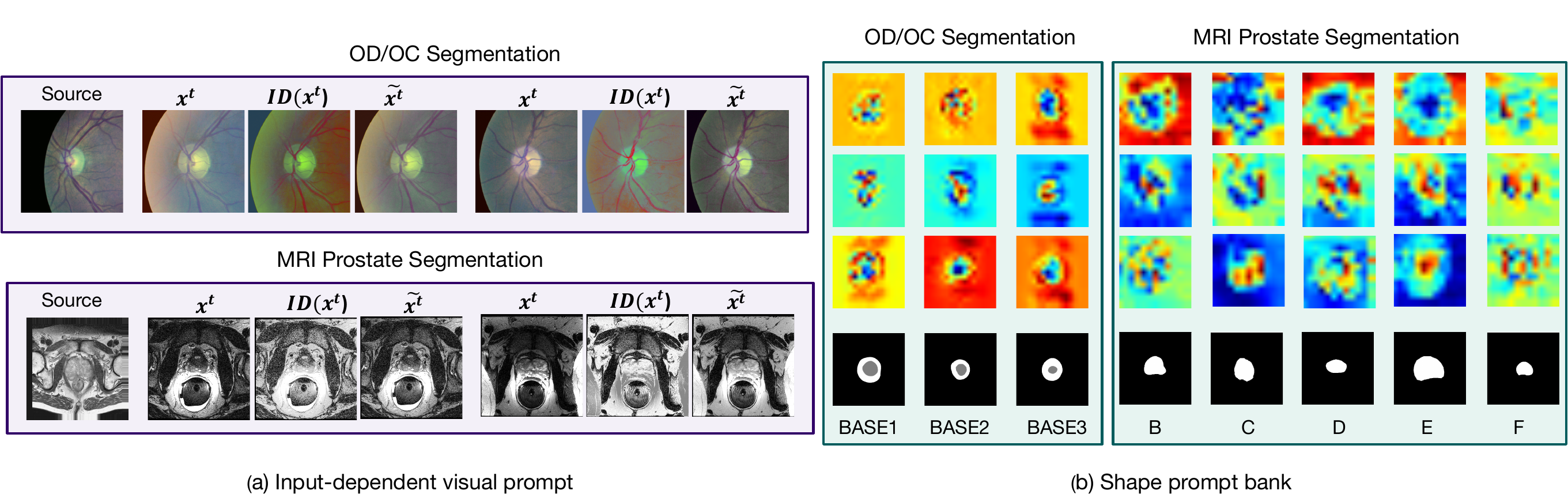}
    \caption{Visualization of (a) the input-dependent visual prompts and (b) the shape prompts in $\overline{SP}$ learned from different unseen target test sets with $L=512$, as well as the corresponding segmentation mask examples.}
    \label{fig::visual_sp}
\end{figure*}
\begin{figure}[!]
    \centering
    \includegraphics[width=1\linewidth]{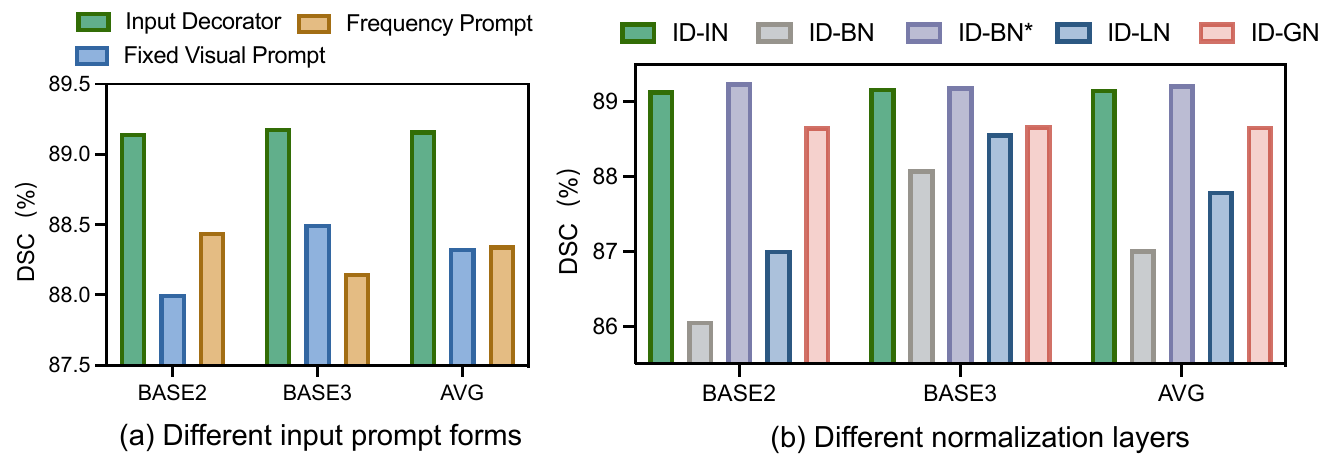}
    \caption{(a) The effect of different input-space prompt formulations on the OD/OC segmentation task for online TTA. Here, we compare the traditional fixed-prompt\cite{hu2022prosfda}, the frequency-domain prompt\cite{chen2023vptta} and the proposed input decorator (ID) that generates input-dependent prompt on the OD/OC segmentation task. (b) The effect of different normalization layers in the proposed ID. IN\cite{IN} / BN\cite{BN} / LN\cite{LN} / GN\cite{GN} stands for Instance/Batch/Layer/Group Normalization, respectively. BN$^*$ means that using BN without tracking the moving average statistics. }
    \label{fig::ablation_input_prompt}
\end{figure}
\begin{figure}[h]
    \centering
    \includegraphics[width=1\linewidth]{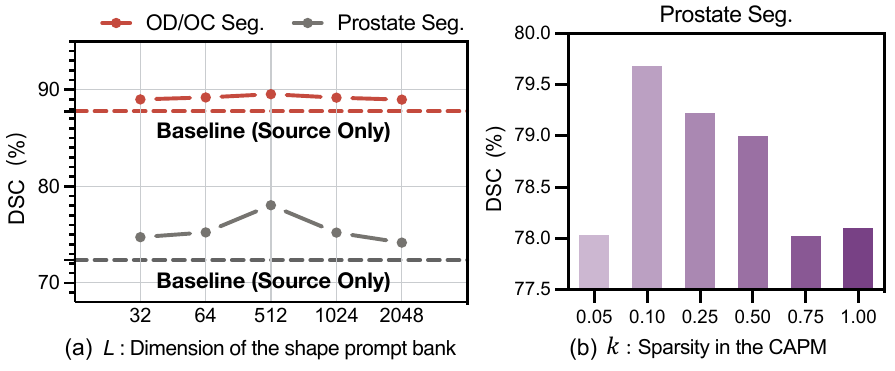}
    \caption{(a) Performance of our PASS with various $L$ on the OD/OC and prostate segmentation tasks for online TTA. (b) Performance of our PASS with various $k$ on the prostate segmentation task for offline TTA.}
    \label{fig::ablation_hyper}
\end{figure}
\begin{figure*}[!ht]
    \centering
    \includegraphics[width=1\linewidth]{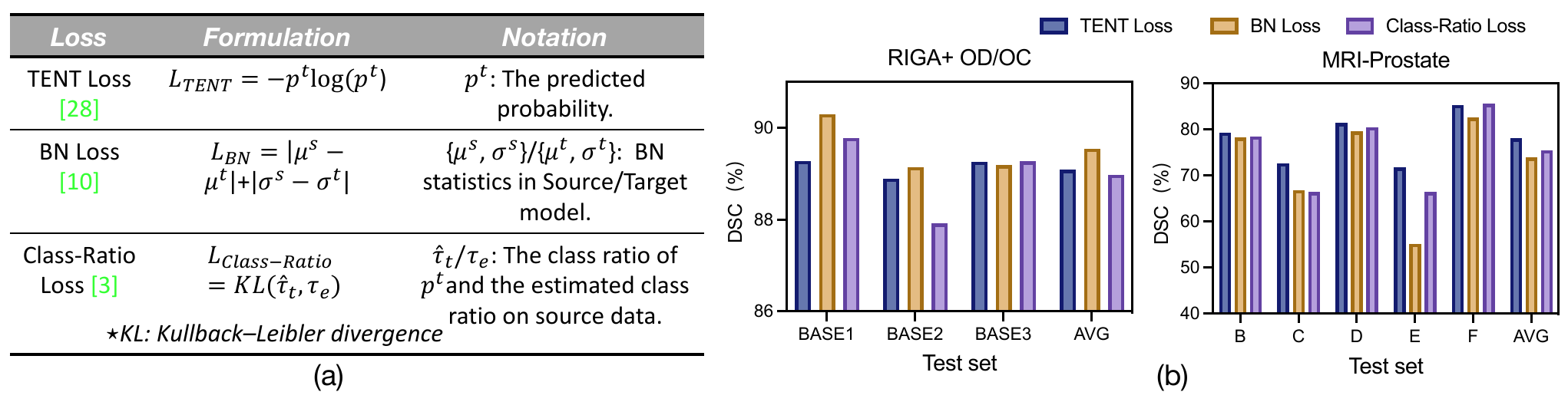}
    \caption{(a) Definition of different losses for TTA. (b) The effect of different loss functions for online TTA on the OD/OC and prostate segmentation tasks.}
    \label{fig::ablation_loss}
\end{figure*}
\begin{figure*}[!t]
    \centering
    \includegraphics[width=0.9\linewidth]{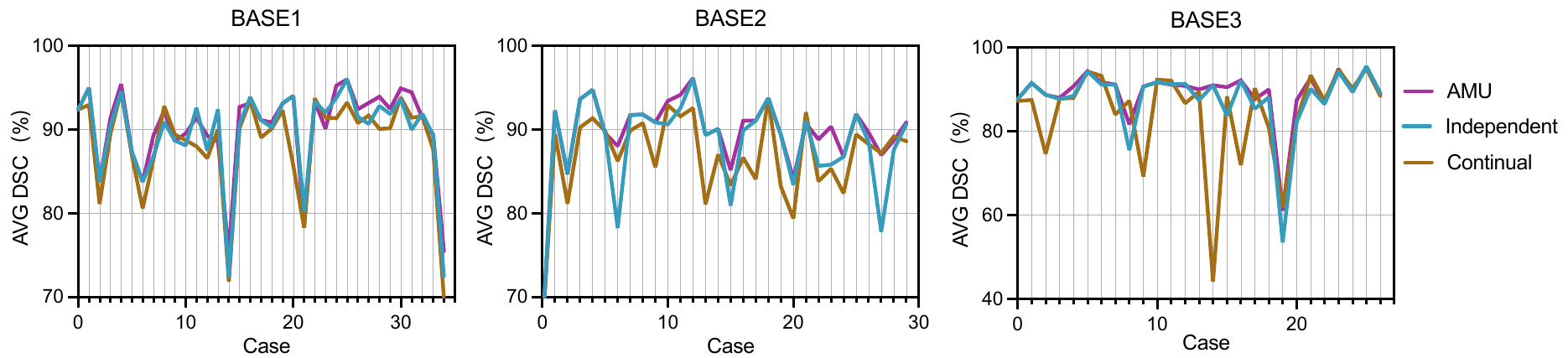}
    \caption{Performance of our PASS with different parameter updating schemes on the OD/OC segmentation task in continual online test-time adaptation. The average dice score of OD and OC classes is reported (AVG DSC).  ``AMU'' stands for our proposed alternating momentum strategy for parameter updating. ``Independent'' represents that adaptation is performed on each test subject independently. ``Continual'' means that the model state for the current sample is initialized with the parameters updated from the previous sample instead of the original source model.}
    \label{fig::continual_RIGA}
\end{figure*}
\begin{figure}[!]
    \centering
    \includegraphics[width=1\linewidth]{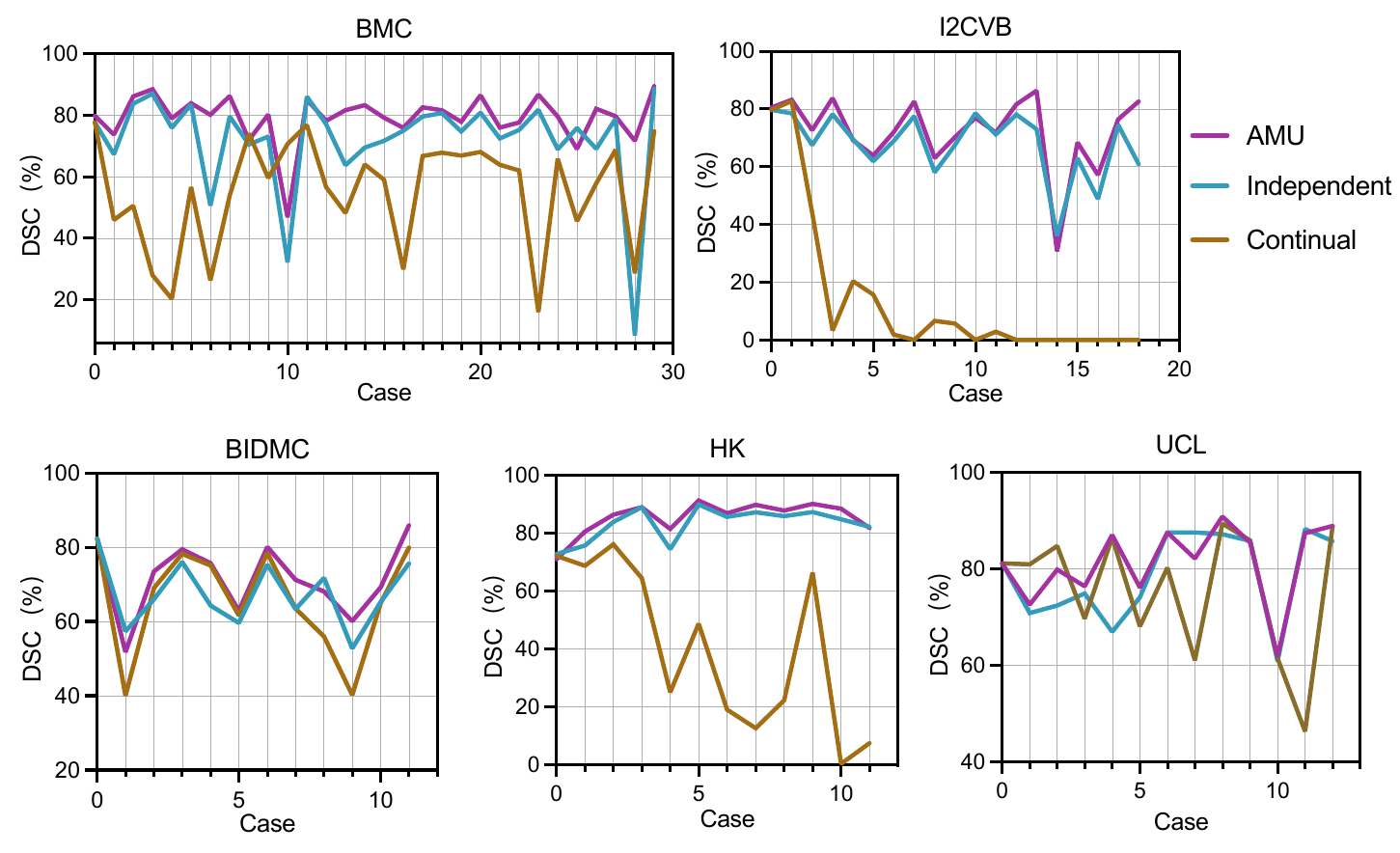}
    \caption{Performance of our PASS with different parameter updating schemes on the MRI prostate segmentation task in continual online test-time adaptation. ``AMU'' stands for our proposed alternating momentum strategy for parameter updating. ``Independent'' represents that adaptation is performed on each test subject independently. ``Continual'' means that the model state for the current sample is initialized with the parameters updated from the previous sample instead of the original source model.}
    \label{fig::continual_prostate}
\end{figure}
\begin{figure}[!ht]
    \centering
    \includegraphics[width=0.8\linewidth]{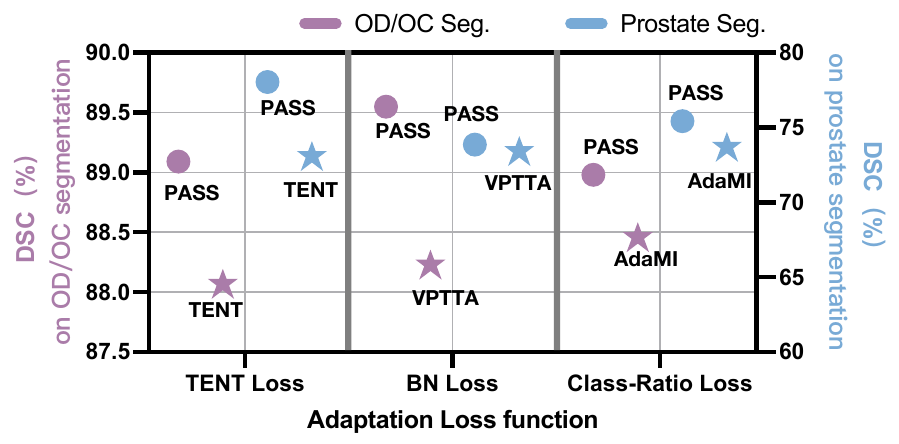}
    \caption{Fair comparison of PASS and other TTA methods with the same adaptation losses.}
    \label{fig::ada_loss_comparison}
\end{figure}
\subsection{Ablation study.}
\indent\textbf{1) Effectiveness of the proposed modules in PASS.} The proposed PASS consists of two prompt modules: an input decorator (ID) and a cross-attention prompt modulator (CAPM). The best performance of PASS over the variants in Table \ref{tab::ablation} demonstrates the contributions of each module. Specifically, not using ID causes a notable performance drop in prostate datasets. This highlights the necessity of decorating the new-coming target test data to eliminate style shifts from source data. In addition, tackling the shape gaps in the semantic representation space through the CAPM also remarkably boosts the generalization of the source model. \changed{M1.1}{\link{R1.1}}{Specifically for the parameter updating scheme in online TTA, we compare the proposed AMU with independent learning \cite{bateson2022moment} (w/o AMU) and the standard EMA strategy\cite{wang2022cotta} (w/ EMA). The results validate that AMU is essential to ensure the stability of online training.}

\textbf{2) The effect of different input-space prompt formulations.} 
To validate the effectiveness of the proposed input-dependent prompt approach over traditional fixed prompts, we replace our ID with the fixed visual prompt used in \cite{hu2022prosfda} and the frequency-domain prompt proposed in \cite{chen2023vptta} on the OD/OC segmentation task. The results in Figure \ref{fig::ablation_input_prompt} (a) prove the better adaptation performance of input-dependent prompt generated by ID than other fixed prompts. \changed{M2.3}{\link{R2.3}}{In Figure \ref{fig::ablation_input_prompt}(b), we explore the effect of normalization variants in ID, including IN\cite{IN}, BN\cite{BN}, LN\cite{LN} and GN\cite{GN}. 
For the online TTA with a batch size of 1, the only difference between BN and IN is that BN normalizes features using historical moving averaged statistics, which is detrimental to preserving the sample-specific style and thus leads to a significant performance drop. If we use BN without tracking the moving average statistics (BN$^*$), its performance is almost identical to IN. Compared with IN and BN which preserve the inter-channel distribution differences, LN eliminates these differences, and GN is an intermediate version of these two types. The performance trend of IN $>$ GN $>$ LN indicates that treating different channels distinctly is necessary for generating effective input prompts.}

\textbf{3) The effect of different adaptation losses.} The proposed PASS is orthogonal to existing TTA methods dedicated to designing effective loss functions and can be integrated with them. We evaluate three objective functions on both the RIGA+ and prostate datasets: the TENT loss in TENT\cite{wang2021tent}, the BN loss in VPTTA\cite{hu2022prosfda} and the Class-Ratio loss in AdaMI\cite{bateson2022moment}. 
The detailed formulations are summarized in Figure \ref{fig::ablation_loss} (a). The online TTA results in Figure \ref{fig::ablation_loss} (b) indicate that the BN and TENT losses perform best in OD/OC and prostate segmentation tasks, respectively. 
Here, we provide suggestions for selecting appropriate TTA loss functions: (1) Overall, TENT loss is a robust choice. (2) BN loss might lead to better performance when the domain gap is minimal. (3) The Class-Ratio loss lacks general applicability since it necessitates the acquisition of source labels.
Additionally, we provide a fair comparative analysis of PASS employing the TNET, BN, and Class-Ratio loss functions against TENT, VPTTA, and AdaMI, in Figure \ref{fig::ada_loss_comparison}
It can be observed that PASS consistently outperforms the other methods when using the same loss, which further demonstrates the effectiveness of the proposed TTA framework.

% To further investigate how to select proper adaptation losses, we compare PASS with the previous TTA methods TENT/VPTTA/AdaMI using the TNET/BN/Class-Ratio losses respectively in Figure \ref{fig::ada_loss_comparison}. Overall, TENT loss is a robust choice when facing unknown test domains while BN loss might lead to better performance when the domain gap is minimal.}

%TENT loss $L_{TENT}=-{\hat{y_T}}\log(\hat{y_T})$\cite{wang2021tent}, the BN loss $L_{BN}=|\mu_S-\mu_T|+|\sigma_S-\sigma_T|$\cite{hu2022prosfda} and the Class-Ratio loss $L_{Class-Ratio}=-{KL(\hat{\tau}_T}, \tau_e)$\cite{bateson2022moment}, in which $\hat{y_T}$ is the predicted probability, ${\mu_S, \sigma_S}/{\mu_T, \sigma_T}$ refer to the BN statistics in the source/target model and $\hat{\tau}_T$/$\tau_e$ stand for the class ratio of $\hat{y_T}$ and the estimated class ratio on source data.
% \begin{figure*}[!]
%     \centering
%     \includegraphics[width=0.7\linewidth]{Fig/ablation_hyperparam.pdf}
%     \caption{(a) The effect of different $L$ which is the dimension of the shape prompt bank $\tilde{SP}$ for online TTA on the prostate datasets. (b) The effect of different decay factors in AMU strategy for online TTA on the OD/OC and prostate segmentation tasks.}
%     \label{fig::ablation_hyperparam}
% \end{figure*}

\textbf{4) The influence of hyperparameters in PASS.}
We study how the size of the shape prompt bank $L$ affects the performance of our method. 
As shown in Figure \ref{fig::ablation_hyper}(a), PASS with a moderate size $L=512$ yields the best average DSC on both applications. \changed{M2.5}{\link{R2.5}}{Intuitively, fewer prompt templates are insufficient to embed the rich style patterns for a target domain, while excessive prompt templates tend to be redundant. Thus, an inappropriate size of $L$ could negatively impact the adaptation stage. Nevertheless, all models with varying $L$ consistently surpass the baseline (Source Only), which proves the robustness of PASS.} Besides, the impact of $L$ on OD/OC segmentation is less than prostate segmentation since the prostate has a more complex shape (as exhibited in Figure \ref{fig::visual_sp}(b)). \changed{M2.7}{\link{R2.7}}{Finally, we conduct the sensitivity analysis regarding $k$ ($top_k$ for the
sparsity of $A$ in CAPM). As illustrated in Figure~\ref{fig::ablation_hyper}(b), $k$ yields higher DSC between 0.1 and 0.5.}

\textbf{5) Visualization of input and shape prompts.} Figure \ref{fig::visual_sp} shows input-space visual prompts and the learned shape prompts on different test sets and the corresponding masks. 
We can observe that (1) the visual prompts alter the texture style of input images. (2) the shape prompts encode rich shape patterns with affinity to the corresponding target objects (the optic disc-cup or prostate). These $L$ shape prompts in $\overline{SP}$ can serve as structure templates and our proposed CAPM extracts pertinent information for each test sample from the extensive shape repository, thereby enhancing the segmentation performance. (3) Given the same source model, each test set learns a distinct shape prompt bank, indicating that different sets possess unique shape prior information. Therefore, it is reasonable to learn individual prompt banks for different target datasets.

% \textcolor{black}

\textbf{6)The influence of optimization schemes in online TTA.} \changed{M2.2}{\link{R2.2}}{Figure \ref{fig::continual_RIGA} and Figure \ref{fig::continual_prostate} compare the online performance of PASS with different parameter updating schemes at sequentially arrived test cases. It can be observed that continual adaptation could cause notable performance degradation in some cases due to potential error accumulation.} For most test datasets, independent adaptation outperforms continual adaptation as the parameter updating for one case does not affect the following cases, which avoids error accumulation. However, the adapted model cannot fully utilise the historical distribution information from former test cases. In contrast, the proposed alternating momentum updating (AMU) strategy leads to more stable adaptation and better performance through independently updating the student network for every single case and recovering knowledge from the teacher network.

\section{Discussion}
 \begin{table}[t!]
\renewcommand{\arraystretch}{0.8} 
\setlength{\extrarowheight}{2pt}
\centering
\caption{Online TTA performance of PASS with different initializations in DSC Score (\%) on three OD/OC segmentation test sets. ``Init.'' stands for initialization. ``Gain'' refers to the magnitude of random distributions. }
\setlength{\tabcolsep}{4pt} % 调整列间距
%>{\columncolor{gray!20}}
% \resizebox{1\textwidth}{!}{
\begin{tabular}{clc}
\toprule
Gain &
{Method} & {Average DSC} \\ \midrule $\sqrt{5}$ &
Kaiming Init.\cite{he2015delving} (default) &  {89.55\scriptsize{±5.49}}        \\ \midrule \multirow{5}{*}{$0\sim0.1$} &
Kaiming Init.\cite{he2015delving} & \textbf{89.63\scriptsize{±5.82}}        \\

&
 Uniform Init.       & 89.24\scriptsize{±5.86} 
\\ &
Xavier Init.\cite{glorot2010understanding}   & 89.26\scriptsize{±5.85} 
\\ &
 Orthogonal Init.\cite{saxe2013exact}  & 88.91\scriptsize{±6.09} \\
\midrule \multirow{5}{*}{$1\sim10$} &
Kaiming Init.\cite{he2015delving}  & {89.14\scriptsize{±6.21}}        \\

&
 Uniform Init.       & 87.38\scriptsize{±5.65} 
\\ &
Xavier Init.\cite{glorot2010understanding}    &  88.62\scriptsize{±6.29} 
\\ &
 Orthogonal Init.\cite{saxe2013exact}   & 87.28\scriptsize{±6.22}
\\ \bottomrule
\end{tabular}
\label{tab::init_type}  
\end{table}

 \begin{figure}[!t]
    \centering
    \includegraphics[width=0.8\linewidth]{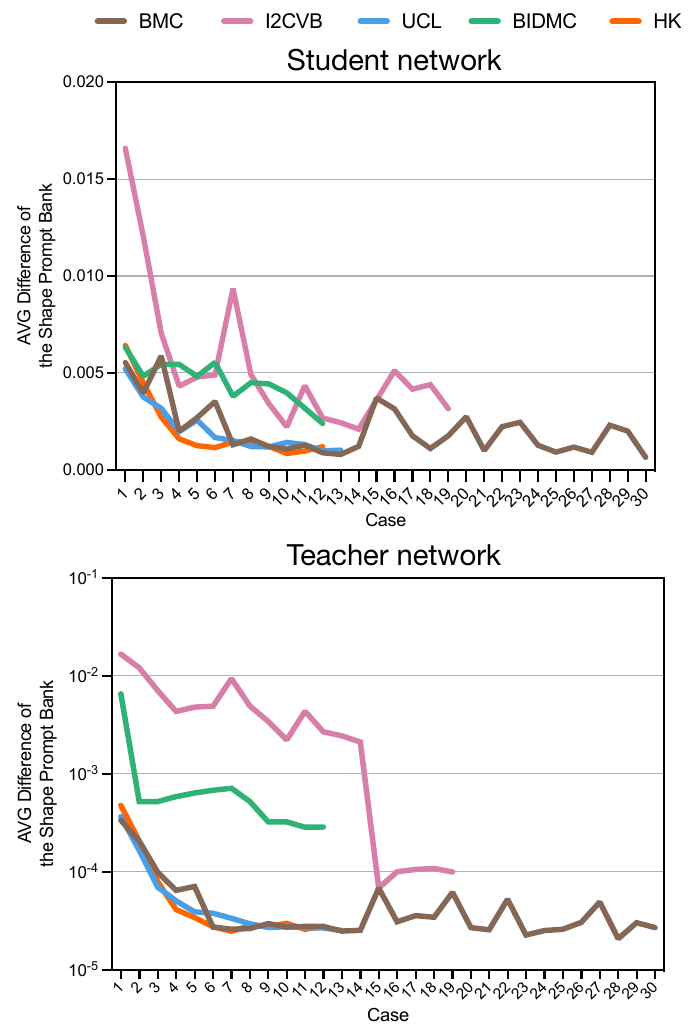}
    \caption{The average difference of the shape prompt bank between adjacent adaptation steps over test cases.}
    \label{fig::shape_prompt_converge}
    \end{figure}
    
\subsection{The effect of initialization}
\changed{M2.1}{\link{R2.1}}{Following most adapter/prompt-based methods when integrating extra parameters into the pretrained backbone\cite{liu2023pre}, both the Input Decorator and CAPM are introduced in a residual manner. This mechanism ensures that the pretrained features remain intact. To validate that PASS is robust to random initialization types, we evaluated five different types of random initialization in Table~\ref{response_tab:init_type}. 
We adjust the magnitude of random distributions across different intervals and present the average performance for a small range $[0.001, 0.005, 0.01, 0.05, 0.1]$ ($0
 \sim0.1$) and a large range $[1, 5, 10]$ ($1
 \sim10$). The magnitude (also termed gain) determines the scale of the random parameter. It should be noted that random initialization magnitudes are generally kept within the range of 0 to 1 to ensure numerical stability during initial gradient propagation. The larger range used here is solely for experimental investigations. We can observe that PASS achieves good performance across different random initializations within the small range. However, in the large range, the introduced random parameters potentially disrupt the pretrained features and cause a significant performance drop. The default Kaiming initialization is robust to varying scales, while orthogonal and uniform initializations exhibit greater sensitivity.
 Overall, within a reasonable range, random initialization would not affect the effectiveness of the proposed method.}

\subsection{Convergence of the CAPM}
\changed{M3.6}{\link{R3.6}}{Figure~\ref{fig::shape_prompt_converge} records the average difference of the shape prompt bank between tuning stages $t$ and $t+1$ in the student network and teacher network. The general trend is that the shape bank will converge in later adaptation steps, and the difference is greater in the student network than in the teacher network. After nearly 6 adaptation steps, the shape bank has learned enriched representations and stabilizes. Since we employ a moving average scheme (i.e., AMU) to update the teacher network, its difference becomes really small ($<1 \times 10^{-4}$) rapidly.}

\begin{figure*}[!]
    \centering
    \includegraphics[width=1\linewidth]{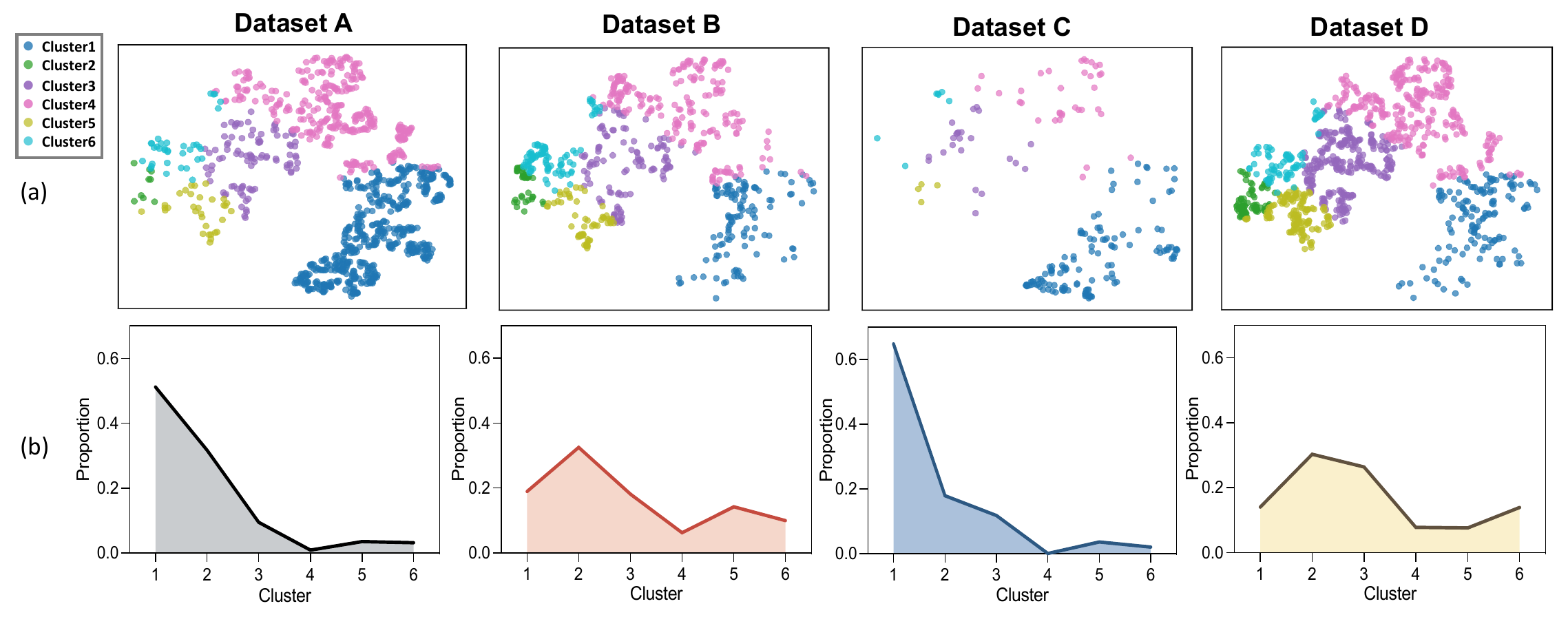}
    \caption{(a) t-SNE visualization of K-means clusters of the shape descriptors in four polyp datasets A-D. (b) The proportions of different clusters.}
    \label{fig::polyp_cluster}
\end{figure*}
\subsection{Extension and limitations}
\label{sec::limitation}
\changed{M3.1.1}{\link{R3.1}}{
To further discuss the outcomes of our method in segmentation tasks with versatile shapes, we extend PASS to a polyp segmentation task. The polyp segmentation task includes four domains\cite{chen2023vptta}: BKAI (A),  CVC (B), ETIS (C), and Kvasir (D), with 1000, 612, 196, and 1000 images, respectively. Each image was resized to 352$\times$352 pixels and normalized with ImageNet statistics. Each dataset is alternately used as the source domain, with the remaining three serving as target domains. Note that all implementation details of the polyp segmentation task are based on the source code of \cite{chen2023vptta}, where the PraNet \cite{fan2020pranet} is used as the backbone.}

\begin{figure}[t]
    %\counterwithin{figure}{section}
    \centering
    \includegraphics[width=1\linewidth]{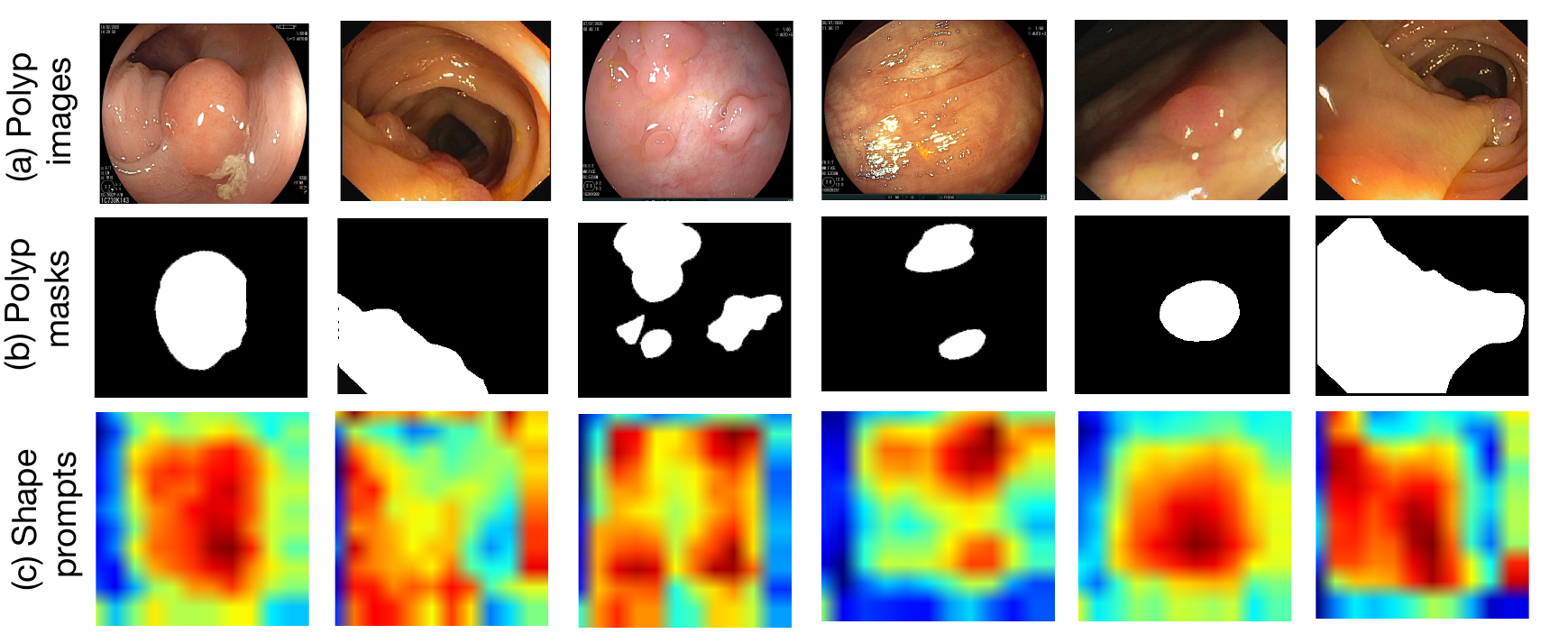}
    \caption{The input images, segmentation masks and learnt shape prompts in the polyp datasets. Each column displays a sample and the relevant shape prompt.}
    \label{fig::polyp_sp}
    \end{figure}

\changed{M3.1.2}{\link{R3.1}}{
The results are shown in Table~\ref{tab::polyp_comparison}. 
We observe that comparable performance to VPTTA\cite{chen2023vptta} can be achieved using only the ID. For the CAPM, although it performs better on domain A/C, a significant performance drop occurs on the other two source domains. To explore the reason, we collect the shape description vectors of the four domains proposed in\cite{bateson2022moment} and perform clustering. The t-SNE visualization in Figure~\ref{fig::polyp_cluster} indicates that the shape distributions of the domain B/D are particularly discrete.  Overall, CAPM excels in models pretrained on data with concentrated shape distributions but shows limited effectiveness with more scattered shape distributions.  In Figure~\ref{fig::polyp_sp}, we find that shape prompt can still provide some templates harbouring the location relationships of objects.
Improvements to its single interaction pattern could be a solution to this limitation, which we leave to our future work.}
 \begin{table}[]
 \renewcommand{\arraystretch}{1} 
 \centering
 \caption{Online TTA performance of different methods in DSC score (\%) on the Polyp segmentation tasks. The comparison results are drawn from \cite{chen2023vptta}. The source model is pretrained on each single domain (source domain) and tested on the left domains (target domains).}
 \setlength{\extrarowheight}{2pt}
\begin{tabular}{l|cccc|c}
\toprule
Method      & A                     & B                   & C                    & D                   & Average                     \\ \hline
Source Only & 79.9                      & 66.33                     & 73.89                     & 82.95                      & 75.77                     \\
PTBN \cite{nado2020ptbn}       & 76.13                     & 75.63                     & 72.28                     & 82.3                       & 76.59                     \\
TENT \cite{wang2021tent}       & 74.86                     & 67.51                     & 17.79                     & 73.55                      & 58.43                     \\
CoTTA \cite{wang2022cotta}      & 76.46                     & 66.77                     & 71.39                     & 70.71                      & 71.33                     \\
DUA \cite{mirza2022dua}        & 78.93                     & 66.84                     & 76.53                     & 86.24                      & 77.13                     \\
SRA  \cite{niu2023sar}       & 76.48                     & 66.45                     & 71.46                     & 70.41                      & 71.20                      \\
VPTTA \cite{chen2023vptta}       & 81.00                     & \underline{76.87}                     & 77.58                     & \underline{86.39}                      & {80.46}                     \\ \hline
PASS      & \textbf{83.12}                   & 76.07
                   & \textbf{77.93}            & 86.01
                    & \textbf{80.78}
                    \\
CAPM Only & \underline{82.13}            & \textcolor{red}{74.53}                     & {77.26}            & \textcolor{red}{82.53}                     & 79.11                   \\
ID  Only& 80.78 & \textbf{77.15} &  \underline{77.61} &   \textbf{86.55}  & \underline{80.52} \\ \bottomrule
\end{tabular}
\label{tab::polyp_comparison}  
\end{table}
    
\section{Conclusion}
In this paper, we have investigated a novel test-time adaptation method named PASS to learn hybrid prompts adapted to the test domain for pretrained segmentation models. Specifically, we proposed an input decorator to alter the low-level style information within input data and a cross-attention prompt modulator, which customizes target-specific shape prompts from diverse latent representations, to enhance the high-level shape prior. An alternating momentum updating mechanism was designed to train the model with continual online test data, preserving historical knowledge and attenuating the impact of newcomers on the model. Without the requirement to access the source domain data and any constraints on the target organ shape, PASS can be easily deployed on the test-time of segmentation models. Extensive experiments have been conducted over various medical imaging datasets to confirm the efficacy and superior performance of PASS. \changed{M3.1.3}{\link{R3.1}}{In the future, we will attempt to propose dynamic shape prompt schemes to tackle the tasks with diverse shape priors.}
\bibliographystyle{IEEEtran}
% \bibliography{reference}

\begin{thebibliography}{10}
\providecommand{\url}[1]{#1}
\csname url@samestyle\endcsname
\providecommand{\newblock}{\relax}
\providecommand{\bibinfo}[2]{#2}
\providecommand{\BIBentrySTDinterwordspacing}{\spaceskip=0pt\relax}
\providecommand{\BIBentryALTinterwordstretchfactor}{4}
\providecommand{\BIBentryALTinterwordspacing}{\spaceskip=\fontdimen2\font plus
\BIBentryALTinterwordstretchfactor\fontdimen3\font minus
  \fontdimen4\font\relax}
\providecommand{\BIBforeignlanguage}[2]{{%
\expandafter\ifx\csname l@#1\endcsname\relax
\typeout{** WARNING: IEEEtran.bst: No hyphenation pattern has been}%
\typeout{** loaded for the language `#1'. Using the pattern for}%
\typeout{** the default language instead.}%
\else
\language=\csname l@#1\endcsname
\fi
#2}}
\providecommand{\BIBdecl}{\relax}
\BIBdecl

\bibitem{zhao2020review}
S.~Zhao, X.~Yue, S.~Zhang, B.~Li, H.~Zhao, B.~Wu, R.~Krishna, J.~E. Gonzalez,
  A.~L. Sangiovanni-Vincentelli, S.~A. Seshia \emph{et~al.}, ``A review of
  single-source deep unsupervised visual domain adaptation,'' \emph{IEEE
  Transactions on Neural Networks and Learning Systems}, vol.~33, no.~2, pp.
  473--493, 2020.

\bibitem{wu2021unsupervised}
F.~Wu and X.~Zhuang, ``Unsupervised domain adaptation with variational
  approximation for cardiac segmentation,'' \emph{IEEE Transactions on Medical
  Imaging}, vol.~40, no.~12, pp. 3555--3567, 2021.

\bibitem{bateson2022source}
M.~Bateson, H.~Kervadec, J.~Dolz, H.~Lombaert, and I.~B. Ayed, ``Source-free
  domain adaptation for image segmentation,'' \emph{Medical Image Analysis},
  vol.~82, p. 102617, 2022.

\bibitem{liu2020shape}
Q.~Liu, Q.~Dou, and P.-A. Heng, ``Shape-aware meta-learning for generalizing
  prostate mri segmentation to unseen domains,'' in \emph{Medical Image
  Computing and Computer Assisted Intervention--MICCAI 2020: 23rd International
  Conference, Lima, Peru, October 4--8, 2020, Proceedings, Part II 23}.\hskip
  1em plus 0.5em minus 0.4em\relax Springer, 2020, pp. 475--485.

\bibitem{wu2023upl}
J.~Wu, G.~Wang, R.~Gu, T.~Lu, Y.~Chen, W.~Zhu, T.~Vercauteren, S.~Ourselin, and
  S.~Zhang, ``Upl-sfda: Uncertainty-aware pseudo label guided source-free
  domain adaptation for medical image segmentation,'' \emph{IEEE transactions
  on medical imaging}, 2023.

\bibitem{wang2021tent}
D.~Wang, E.~Shelhamer, S.~Liu, B.~Olshausen, and T.~Darrell, ``Tent: Fully
  test-time adaptation by entropy minimization,'' in \emph{International
  Conference on Learning Representations}, 2021, p. 9229–9248.

\bibitem{wang2022cotta}
Q.~Wang, O.~Fink, L.~Van~Gool, and D.~Dai, ``Continual test-time domain
  adaptation,'' in \emph{Proceedings of the IEEE/CVF Conference on Computer
  Vision and Pattern Recognition}, 2022, pp. 7201--7211.

\bibitem{bateson2022moment}
M.~Bateson, H.~Lombaert, and I.~Ben~Ayed, ``Test-time adaptation with shape
  moments for image segmentation,'' in \emph{International Conference on
  Medical Image Computing and Computer-Assisted Intervention}.\hskip 1em plus
  0.5em minus 0.4em\relax Springer, 2022, pp. 736--745.

\bibitem{he2020self}
Y.~He, A.~Carass, L.~Zuo, B.~E. Dewey, and J.~L. Prince, ``Self domain adapted
  network,'' in \emph{Medical Image Computing and Computer Assisted
  Intervention--MICCAI 2020: 23rd International Conference, Lima, Peru, October
  4--8, 2020, Proceedings, Part I 23}.\hskip 1em plus 0.5em minus 0.4em\relax
  Springer, 2020, pp. 437--446.

\bibitem{niu2023sar}
S.~Niu, J.~Wu, Y.~Zhang, Z.~Wen, Y.~Chen, P.~Zhao, and M.~Tan, ``Towards stable
  test-time adaptation in dynamic wild world,'' in \emph{International
  Conference on Learning Representations}, 2023.

\bibitem{sun2020test}
Y.~Sun, X.~Wang, Z.~Liu, J.~Miller, A.~Efros, and M.~Hardt, ``Test-time
  training with self-supervision for generalization under distribution
  shifts,'' in \emph{International Conference on Machine Learning}.\hskip 1em
  plus 0.5em minus 0.4em\relax PMLR, 2020, pp. 9229--9248.

\bibitem{valanarasu2022dpg}
J.~M.~J. Valanarasu, P.~Guo, V.~VS, and V.~M. Patel, ``On-the-fly test-time
  adaptation for medical image segmentation,'' \emph{arXiv preprint
  arXiv:2203.05574}, 2022.

\bibitem{he2021autoencoder}
Y.~He, A.~Carass, L.~Zuo, B.~E. Dewey, and J.~L. Prince, ``Autoencoder based
  self-supervised test-time adaptation for medical image analysis,''
  \emph{Medical image analysis}, vol.~72, p. 102136, 2021.

\bibitem{guo2023atals}
J.~Guo, W.~Zhang, M.~Sinclair, D.~Rueckert, and C.~Chen, ``Pay attention to the
  atlas: Atlas-guided test-time adaptation method for robust 3d medical image
  segmentation,'' \emph{arXiv preprint arXiv:2307.00676}, 2023.

\bibitem{hu2021rncr}
M.~Hu, T.~Song, Y.~Gu, X.~Luo, J.~Chen, Y.~Chen, Y.~Zhang, and S.~Zhang,
  ``Fully test-time adaptation for image segmentation,'' in \emph{Medical Image
  Computing and Computer Assisted Intervention--MICCAI 2021: 24th International
  Conference, Strasbourg, France, September 27--October 1, 2021, Proceedings,
  Part III 24}.\hskip 1em plus 0.5em minus 0.4em\relax Springer, 2021, pp.
  251--260.

\bibitem{wu2023upltta}
J.~Wu, R.~Gu, T.~Lu, S.~Zhang, and G.~Wang, ``Upl-tta: Uncertainty-aware pseudo
  label guided fully test time adaptation for fetal brain segmentation,'' in
  \emph{International Conference on Information Processing in Medical
  Imaging}.\hskip 1em plus 0.5em minus 0.4em\relax Springer, 2023, pp.
  237--249.

\bibitem{nguyen2023tipi}
A.~T. Nguyen, T.~Nguyen-Tang, S.-N. Lim, and P.~H. Torr, ``Tipi: Test time
  adaptation with transformation invariance,'' in \emph{Proceedings of the
  IEEE/CVF Conference on Computer Vision and Pattern Recognition}, 2023, pp.
  24\,162--24\,171.

\bibitem{zhang2023oclttt}
Y.~Zhang, Y.~Sun, S.~Zheng, Z.~Shui, C.~Zhu, and L.~Yang, ``Test-time training
  for semantic segmentation with output contrastive loss,'' \emph{arXiv
  preprint arXiv:2311.07877}, 2023.

\bibitem{nado2020ptbn}
Z.~Nado, S.~Padhy, D.~Sculley, A.~D'Amour, B.~Lakshminarayanan, and J.~Snoek,
  ``Evaluating prediction-time batch normalization for robustness under
  covariate shift,'' \emph{arXiv preprint arXiv:2006.10963}, 2020.

\bibitem{mirza2022dua}
M.~J. Mirza, J.~Micorek, H.~Possegger, and H.~Bischof, ``The norm must go on:
  Dynamic unsupervised domain adaptation by normalization,'' in
  \emph{Proceedings of the IEEE/CVF Conference on Computer Vision and Pattern
  Recognition}, 2022, pp. 14\,765--14\,775.

\bibitem{liu2022single}
Q.~Liu, C.~Chen, Q.~Dou, and P.-A. Heng, ``Single-domain generalization in
  medical image segmentation via test-time adaptation from shape dictionary,''
  in \emph{Proceedings of the AAAI Conference on Artificial Intelligence},
  vol.~36, no.~2, 2022, pp. 1756--1764.

\bibitem{jia2022visual}
M.~Jia, L.~Tang, B.-C. Chen, C.~Cardie, S.~Belongie, B.~Hariharan, and S.-N.
  Lim, ``Visual prompt tuning,'' in \emph{European Conference on Computer
  Vision}.\hskip 1em plus 0.5em minus 0.4em\relax Springer, 2022, pp. 709--727.

\bibitem{zhou2022coop}
K.~Zhou, J.~Yang, C.~C. Loy, and Z.~Liu, ``Learning to prompt for
  vision-language models,'' \emph{International Journal of Computer Vision},
  vol. 130, no.~9, pp. 2337--2348, 2022.

\bibitem{shu2022test}
M.~Shu, W.~Nie, D.-A. Huang, Z.~Yu, T.~Goldstein, A.~Anandkumar, and C.~Xiao,
  ``Test-time prompt tuning for zero-shot generalization in vision-language
  models,'' \emph{Advances in Neural Information Processing Systems}, vol.~35,
  pp. 14\,274--14\,289, 2022.

\bibitem{hu2022prosfda}
S.~Hu, Z.~Liao, and Y.~Xia, ``Prosfda: Prompt learning based source-free domain
  adaptation for medical image segmentation,'' \emph{arXiv preprint
  arXiv:2211.11514}, 2022.

\bibitem{chen2023vptta}
Z.~Chen, Y.~Ye, M.~Lu, Y.~Pan, and Y.~Xia, ``Each test image deserves a
  specific prompt: Continual test-time adaptation for 2d medical image
  segmentation,'' \emph{arXiv preprint arXiv:2311.18363}, 2023.

\bibitem{karani2021dae}
N.~Karani, E.~Erdil, K.~Chaitanya, and E.~Konukoglu, ``Test-time adaptable
  neural networks for robust medical image segmentation,'' \emph{Medical Image
  Analysis}, vol.~68, p. 101907, 2021.

\bibitem{liu2023pre}
P.~Liu, W.~Yuan, J.~Fu, Z.~Jiang, H.~Hayashi, and G.~Neubig, ``Pre-train,
  prompt, and predict: A systematic survey of prompting methods in natural
  language processing,'' \emph{ACM Computing Surveys}, vol.~55, no.~9, pp.
  1--35, 2023.

\bibitem{loedeman2022prompt}
J.~Loedeman, M.~C. Stol, T.~Han, and Y.~M. Asano, ``Prompt generation networks
  for efficient adaptation of frozen vision transformers,'' \emph{arXiv
  preprint arXiv:2210.06466}, 2022.

\bibitem{nie2023pro}
X.~Nie, B.~Ni, J.~Chang, G.~Meng, C.~Huo, S.~Xiang, and Q.~Tian, ``Pro-tuning:
  Unified prompt tuning for vision tasks,'' \emph{IEEE Transactions on Circuits
  and Systems for Video Technology}, 2023.

\bibitem{gan2023decorate}
Y.~Gan, Y.~Bai, Y.~Lou, X.~Ma, R.~Zhang, N.~Shi, and L.~Luo, ``Decorate the
  newcomers: Visual domain prompt for continual test time adaptation,'' in
  \emph{Proceedings of the AAAI Conference on Artificial Intelligence},
  vol.~37, no.~6, 2023, pp. 7595--7603.

\bibitem{huang2017arbitrary}
X.~Huang and S.~Belongie, ``Arbitrary style transfer in real-time with adaptive
  instance normalization,'' in \emph{Proceedings of the IEEE international
  conference on computer vision}, 2017, pp. 1501--1510.

\bibitem{almazroa2018retinal}
A.~Almazroa, S.~Alodhayb, E.~Osman, E.~Ramadan, M.~Hummadi, M.~Dlaim,
  M.~Alkatee, K.~Raahemifar, and V.~Lakshminarayanan, ``Retinal fundus images
  for glaucoma analysis: the riga dataset,'' in \emph{Medical Imaging 2018:
  Imaging Informatics for Healthcare, Research, and Applications}, vol.
  10579.\hskip 1em plus 0.5em minus 0.4em\relax SPIE, 2018, pp. 55--62.

\bibitem{decenciere2014feedback}
E.~Decenci{\`e}re, X.~Zhang, G.~Cazuguel, B.~Lay, B.~Cochener, C.~Trone,
  P.~Gain, R.~Ordonez, P.~Massin, A.~Erginay \emph{et~al.}, ``Feedback on a
  publicly distributed image database: the messidor database,'' \emph{Image
  Analysis \& Stereology}, vol.~33, no.~3, pp. 231--234, 2014.

\bibitem{nicholas2015nci}
B.~Nicholas, M.~Anant, H.~Henkjan, F.~John, K.~Justin \emph{et~al.},
  ``Nci-proc. ieee-isbi conf. 2013 challenge: Automated segmentation of
  prostate structures,'' \emph{The Cancer Imaging Archive}, vol.~5, 2015.

\bibitem{litjens2014evaluation}
G.~Litjens, R.~Toth, W.~Van De~Ven, C.~Hoeks, S.~Kerkstra, B.~Van~Ginneken,
  G.~Vincent, G.~Guillard, N.~Birbeck, J.~Zhang \emph{et~al.}, ``Evaluation of
  prostate segmentation algorithms for mri: the promise12 challenge,''
  \emph{Medical image analysis}, vol.~18, no.~2, pp. 359--373, 2014.

\bibitem{lemaitre2015computer}
G.~Lema{\^\i}tre, R.~Mart{\'\i}, J.~Freixenet, J.~C. Vilanova, P.~M. Walker,
  and F.~Meriaudeau, ``Computer-aided detection and diagnosis for prostate
  cancer based on mono and multi-parametric mri: a review,'' \emph{Computers in
  biology and medicine}, vol.~60, pp. 8--31, 2015.

\bibitem{ronneberger2015u}
O.~Ronneberger, P.~Fischer, and T.~Brox, ``U-net: Convolutional networks for
  biomedical image segmentation,'' in \emph{Medical Image Computing and
  Computer-Assisted Intervention--MICCAI 2015: 18th International Conference,
  Munich, Germany, October 5-9, 2015, Proceedings, Part III 18}.\hskip 1em plus
  0.5em minus 0.4em\relax Springer, 2015, pp. 234--241.

\bibitem{yang2022dltta}
H.~Yang, C.~Chen, M.~Jiang, Q.~Liu, J.~Cao, P.~A. Heng, and Q.~Dou, ``Dltta:
  Dynamic learning rate for test-time adaptation on cross-domain medical
  images,'' \emph{IEEE Transactions on Medical Imaging}, vol.~41, no.~12, pp.
  3575--3586, 2022.

\bibitem{IN}
D.~Ulyanov, A.~Vedaldi, and V.~Lempitsky, ``Improved texture networks:
  Maximizing quality and diversity in feed-forward stylization and texture
  synthesis,'' in \emph{Proceedings of the IEEE conference on computer vision
  and pattern recognition}, 2017, pp. 6924--6932.

\bibitem{BN}
S.~Ioffe, ``Batch normalization: Accelerating deep network training by reducing
  internal covariate shift,'' \emph{arXiv preprint arXiv:1502.03167}, 2015.

\bibitem{LN}
J.~L. Ba, ``Layer normalization,'' \emph{arXiv preprint arXiv:1607.06450},
  2016.

\bibitem{GN}
Y.~Wu and K.~He, ``Group normalization,'' in \emph{Proceedings of the European
  conference on computer vision (ECCV)}, 2018, pp. 3--19.

\bibitem{he2015delving}
K.~He, X.~Zhang, S.~Ren, and J.~Sun, ``Delving deep into rectifiers: Surpassing
  human-level performance on imagenet classification,'' in \emph{Proceedings of
  the IEEE international conference on computer vision}, 2015, pp. 1026--1034.

\bibitem{glorot2010understanding}
X.~Glorot and Y.~Bengio, ``Understanding the difficulty of training deep
  feedforward neural networks,'' in \emph{Proceedings of the thirteenth
  international conference on artificial intelligence and statistics}.\hskip
  1em plus 0.5em minus 0.4em\relax JMLR Workshop and Conference Proceedings,
  2010, pp. 249--256.

\bibitem{saxe2013exact}
A.~M. Saxe, J.~L. McClelland, and S.~Ganguli, ``Exact solutions to the
  nonlinear dynamics of learning in deep linear neural networks,'' \emph{arXiv
  preprint arXiv:1312.6120}, 2013.

\bibitem{fan2020pranet}
D.-P. Fan, G.-P. Ji, T.~Zhou, G.~Chen, H.~Fu, J.~Shen, and L.~Shao, ``Pranet:
  Parallel reverse attention network for polyp segmentation,'' in
  \emph{International conference on medical image computing and
  computer-assisted intervention}.\hskip 1em plus 0.5em minus 0.4em\relax
  Springer, 2020, pp. 263--273.

\end{thebibliography}

\end{document}